%% file: main.tex
\documentclass{IEEEtaes}

\usepackage{color,array,amsthm}
\usepackage{amsmath}
\usepackage{graphicx}
\usepackage{rotating}
\usepackage{tabularray}
\usepackage{ninecolors}
\usepackage{subcaption}
\usepackage{svg}
\usepackage{amsfonts}
\usepackage{orcidlink}

\jvol{XX}
\jnum{XX}
\jmonth{XXXXX}
\paper{1234567}
\pubyear{2024}
\doiinfo{TAES.2020.Doi Number}

\DeclareMathOperator{\img}{\mathcal{L}_{\text{Img}}}
\DeclareMathOperator{\inst}{\mathcal{L}_{\text{Inst}}}

\DeclareMathOperator{\pc}{\mathcal{L}_{\text{PC}}}
\DeclareMathOperator{\adv}{\mathcal{L}_{\text{Adv}}}

\newcommand{\qualsize}{0.2}

\begin{document}
% \title{You Only Crash Once v2: Domain Adaptive Planetary and Small Body Terrain Detection by Visual Similarities}
% \title{You Only Crash Once v2: Exploiting Feature Similarity for Onboard Domain Adaptive Detection of Space Terrain}
% \title{You Only Crash Once v2: Improving Feature Similarity for Domain Adaptive Detection of Space Terrain}
% \title{You Only Crash Once v2: Exploiting Feature Similarity for Domain Adaptive In-Situ Detection of Space Terrain}
% \title{You Only Crash Once v2: Improving Cross Domain Feature Similarity for Adaptive In-Situ Detection of Space Terrain}
% \title{You Only Crash Once v2: Domain Adaptive Space Terrain Detection by Improved Similarity Alignment}
% \title{You Only Crash Once v2: Intra-Feature Selection and Perceptual Consistency for Domain Adaptive Space Terrain Detection}
% \title{You Only Crash Once v2: Domain Adaptive Object Detection of Space Terrain by Perceptual Consistency and Strong Feature Filtering}
% \title{You Only Crash Once v2: Perceptual Consistency and Strong Feature Filtering for One-Stage Domain Adaptive Detection of Space Terrain}
% \title{You Only Crash Once v2: One-Stage Domain Adaptive Detection of Space Terrain by Perceptually Consistent Strong Features}
\title{You Only Crash Once v2: Perceptually Consistent Strong Features for One-Stage Domain Adaptive Detection of Space Terrain}

\author{Timothy Chase Jr~\orcidlink{0009-0004-3075-0790}}\affil{University at Buffalo, Buffalo, NY 14260, USA}
% \affil{University at Buffalo\\NASA Goddard Space Flight Center} 
% \author{Christopher Gnam}
% \affil{University at Buffalo, Buffalo, NY 14260, USA} 
\author{Christopher Wilson}\affil{NASA Goddard Space Flight Center, Greenbelt, MD 20771, USA} 
\author{Karthik Dantu~\orcidlink{0000-0002-7497-6722}}\affil{University at Buffalo, Buffalo, NY 14260, USA} 

\receiveddate{Manuscript received XXXXX 00, 0000; revised XXXXX 00, 0000; accepted XXXXX 00, 0000.
% This work was supported in part by the U.S. Department of Commerce under Grant BS123456.
}
%% \accepteddate{XXXXX XX XXXX}
%% \publisheddate{XXXXX XX XXXX}

\corresp{Corresponding author: Timothy Chase Jr.}

\authoraddress{Timothy Chase Jr and Karthik Dantu are with the University at Buffalo, Buffalo, NY 14260, USA. Christopher Wilson is with NASA Goddard Space Flight Center, Greenbelt, MD 20771, USA.}

% \editor{Mentions of supplemental materials and animal/human rights statements can be included here.}
% \supplementary{Color versions of one or more of the figures in this article are available online at \href{http://ieeexplore.ieee.org}{http://ieeexplore.ieee.org}.}

\markboth{Chase Jr Et al.}{You Only Crash Once v2}
\maketitle

\begin{abstract}
The in-situ detection of planetary, lunar, and small-body surface terrain is crucial for autonomous spacecraft applications, where learning-based computer vision methods are increasingly employed to enable intelligence without prior information or human intervention. However, many of these methods remain computationally expensive for spacecraft processors and prevent real-time operation. Training of such algorithms is additionally complex due to the scarcity of labeled data and reliance on supervised learning approaches. Unsupervised Domain Adaptation (UDA) offers a promising solution by facilitating model training with disparate data sources such as simulations or synthetic scenes, although UDA is difficult to apply to celestial environments where challenging feature spaces are paramount. To alleviate such issues, You Only Crash Once (YOCOv1) has studied the integration of Visual Similarity-based Alignment (VSA) into lightweight one-stage object detection architectures to improve space terrain UDA. Although proven effective, the approach faces notable limitations, including performance degradations in multi-class and high-altitude scenarios. Building upon the foundation of YOCOv1, we propose novel additions to the VSA scheme that enhance terrain detection capabilities under UDA, and our approach is evaluated across both simulated and real-world data. Our second YOCO rendition, YOCOv2, is capable of achieving state-of-the-art UDA performance on surface terrain detection, where we showcase improvements upwards of 31\% compared with YOCOv1 and terrestrial state-of-the-art. We demonstrate the practical utility of YOCOv2 with spacecraft flight hardware performance benchmarking and qualitative evaluation of NASA mission data. \textcolor{blue}{All code and datasets will be made open source upon publication.}
\end{abstract}

% , an approach traditionally supported by two-stage object detection architectures, by intra-feature clustering

% Traditionally compatible with two-stage object detection architectures,  into more efficient one-stage object detectors that exhibits improved 

% You Only Crash Once (YOCO) integrates Visual Similarity-based Alignment (VSA) by intra-feature clustering into the more efficient one-stage object detection framework of YOLO, promoting improved performance in the presence of space data. 

% \begin{IEEEkeywords}Enter keywords or phrases in alphabetical order, separated by commas.
% \end{IEEEkeywords}

\input{sections/intro}

\input{sections/relwork}
\input{sections/method}
\input{sections/eval}

% \input{sections/discussion}
\input{sections/conclusion}

\clearpage
\bibliographystyle{IEEEtaes.bst}
\bibliography{main.bib}
\end{document}

%% file: sections/intro.tex
\section{Introduction}\label{sec:intro}
\input{figs/intro}

The detection of celestial surface terrain is critical for autonomous spaceflight applications such as Terrain Relative Navigation (TRN), Entry, Descent, and Landing (EDL), hazard analysis, and scientific data collection. Traditional methods are limited by the computational constraints of radiation-hardened systems, relying on extensive \textit{a priori} imaging and offline processing by human operators~\cite{nft1,nft2,retina,m2020lvs}. While effective in missions like the Mars Perseverance Rover landing~\cite{m2020lvs} and OSIRIS-REx's touch-and-go~\cite{nft1,nft2}, these approaches extend mission timelines, increase costs, lack generalization, and execute at low processing rates. The recent advent of space-grade neural network accelerators~\cite{sclearn} enables the deployment of state-of-the-art computer vision techniques, such as object detection, offering real-time performance~\cite{profiling} and robust recognition~\cite{mars} to overcome these limitations.

A major constraint in current object detection methods is their reliance on supervised learning, which requires extensive labeled data to detect and classify objects effectively. This dependency poses significant challenges for spaceflight applications, as celestial surface imagery is limited, and ground truth labels are often difficult or impossible to obtain, especially for missions targeting uncharted environments. In terrestrial vision tasks, Unsupervised Domain Adaptation (UDA) addresses similar issues by training models on accessible, distinct data distributions and adapting them to the target environment. This approach is particularly suited for space applications, where photorealistic simulations or lower-resolution mission data can serve as initial training sources.

Applying UDA to celestial surface imaging is particularly challenging due to difficult illumination conditions, textureless regions, and feature space similarities inherent in unstructured space terrain~\cite{slamprob1,slamprob2,slamprob3,mht}. A promising method to address these challenges is Visual Similarity-based Alignment (VSA). Initially developed for two-stage detectors~\cite{visga}, VSA groups object features by visual appearance into instance-level clusters, reducing distribution shifts without relying on noisy or inaccurate class labels during training. However, two-stage detectors are generally too computationally expensive for spacecraft accelerators, requiring a one-stage solution for real-time operation~\cite{profiling}. Efforts to adapt VSA for one-stage models through the \textit{You Only Crash Once} (YOCO~\cite{yoco}) framework have been largely successful but still face performance degradations in overly smooth or feature-similar terrains, such as those exhibited during high-altitude imagery of the Martian surface.

In this work, we present the next iteration of YOCO tailored to a broader range of one-stage architectures and propose an enhanced VSA formulation for real-time, onboard terrain detection in challenging planetary, lunar, and small-body environments. We target improvements in the presence of textureless regions and varying illumination conditions, integrating recent advances in image generation to mitigate distribution gaps. We investigate a comprehensive set of VSA techniques, including instance- and intra-feature-based clustering within adversarial and contrastive learning frameworks, identifying correlations between clustering methods and terrain types. Through rigorous analysis on simulated and real-world datasets, we achieve a performance improvement of over 31\% compared to YOCOv1 and terrestrial state-of-the-art methods.
Overall, we make the following contributions in this work:
\vspace{-1em}
\begin{itemize}
    \item We perform a deep characterization of the \textit{You Only Look Once} (YOLO) model family for celestial surface terrain detection in UDA settings.
    \item We benchmark \textbf{eight} YOLO model variants on NASA spacecraft hardware, determining the applicability toward in-situ, real-time operation.
    \item We devise a novel formulation of VSA within YOLO, proposing improvements in perceptual feature comparisons and robust feature selection that bolster UDA performance.
    % \item We question the intra-feature clustering used for VSA in prior works, implementing and comparing additional classical and dynamic clustering schemes.
    \item We develop six datasets consisting of both simulated and real-world data emulating Mars, Moon, and Asteroid terrain surfaces for UDA evaluation.
    \item We perform extensive quantitative evaluation, analyzing correlations between the type of VSA and celestial environment where we showcase improved UDA transfer by upwards of 31\%.
    \item We qualitatively analyze our method on real-world mission data.
\end{itemize}

%% file: figs/intro.tex
\begin{figure}[t!]
    \centering
    \includegraphics[width=\columnwidth]{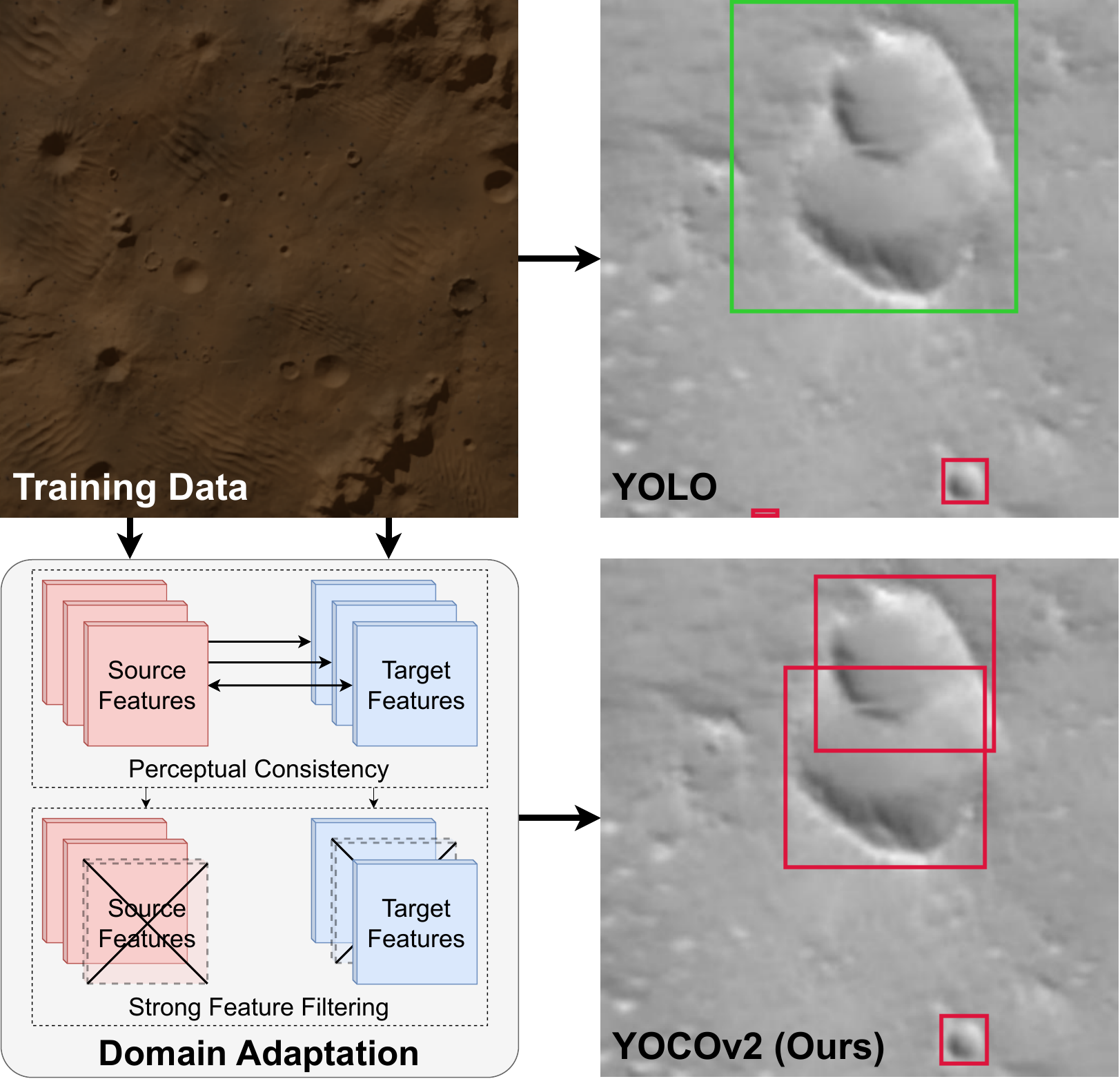}
    \caption{
    Space environments present feature spaces that challenge terrain detection techniques trained with data sources distinct from the operational environment. YOCOv2 introduces a robust feature selection scheme coupled with perceptually consistent regularization that permits performant solutions in lightweight one-stage networks capable of running in real-time on current-generation spacecraft hardware. \textbf{Top Right:} YOLO model with erroneous detections (green, sand). \textbf{Bottom right:} Accurate detections from YOCOv2 (red, crater).
    }\label{fig:enter-label}
    \vspace{-2em}
\end{figure}

%% file: sections/relwork.tex
\section{Related Work}\label{sec:relwork}

\subsection{Space Terrain Detection}
Celestial terrain detection has historically focused on lunar craters, and has evolved from classical computer vision methods to deep learning approaches. Convolutional Neural Networks (CNNs) have been applied for crater classification via edge detection~\cite{cda1} and synthetic data-driven cascaded networks~\cite{cda9}, while U-Net architectures trained on Digital Elevation Maps (DEMs)~\cite{cda2} and infrared maps~\cite{cda3,cda4} have enabled segmentation. The fusion of CNNs, image processing, and DEM data has been additionally successful~\cite{cda8,cda10}. Object detection models like YOLOv3~\cite{cda5}, Faster R-CNN~\cite{cda6}, and Mask R-CNN~\cite{cda7} were first used with curated datasets from previous missions. Limited research has addressed general multi-class terrain detection in varying environments, however, with the only such works employing density analysis for hazard labeling of SIFT features~\cite{terrain1} and ensemble learning with multiple Faster R-CNNs~\cite{terrain3}. The only approach studying UDA, YOCOv1~\cite{yoco}, focused on intra-feature clustering within the YOLOv3 architecture.

These methods face numerous practical challenges. Traditional image processing demands computational optimization~\cite{mht, profiling}, and object detection models like YOLOv3 and Faster R-CNN cannot perform real-time inference on current spacecraft accelerators~\cite{profiling}. Synthetic training data often suffers from domain shift, while domain adaptation remains underexplored in space applications. Prior methods focus only on craters, offering limited generalization to other terrain such as dunes, mountains, and ridges. Multi-class scenarios further complicate dataset creation, where reliance on DEMs or hand labeling is both resource-intensive and non-transferable to novel, unexplored environments; a critical limitation to future missions to worlds such as Titan~\cite{titan} and Enceladus~\cite{enceladus}.

\vspace{-1em}

\input{figs/arch}

\subsection{Visual Similarity-based Domain Adaptation}
VSA was initially introduced within two-stage frameworks that align source and target domain proposals based on visual similarity clustering rather than class or pseudo-class labels~\cite{visga}. Subsequent enhancements include online algorithms~\cite{vsa_two2} and image-to-image translation for intermediate and style-consistent domain data~\cite{vsa_two3}. Grounding pseudo-labels have been adopted~\cite{vsa_two6} as well as mean teacher-student frameworks~\cite{vsa_two9}. Transformers have been employed for improved adaptation~\cite{vsa_transformer1}, while auxiliary heads that compensate for discarded task-relevant features have enhanced performance~\cite{vsa_two10}.

Adopting VSA into one-stage detectors has been challenging due to the lack of disparate region proposals, prompting the exploration of alternative network designs. Single unlabeled target images were utilized~\cite{vsa_yolo1}, along with adversarial alignment combined with self-training to address style gaps~\cite{vsa_yolo3} and employing weak supervision~\cite{vsa_yolo2} or style-invariant losses~\cite{vsa_retinanet}. Instance-level domain alignment has been enhanced using self-attention and novel discriminator architectures~\cite{vsa_one_stage_general} while merging target and source samples~\cite{vsa_yolo7} and employing multi-scale adversarial learning~\cite{vsa_one_stage_general3} have also demonstrated success.

\vspace{-1em}

\subsection{VSA on Celestial Terrain}
Celestial terrain images pose significant challenges due to homogeneous regions, smooth surfaces, and a lack of distinct gradients or textures, especially at higher altitudes. These factors complicate traditional object detection pipelines, particularly in the presence of domain gaps. VSA offers a promising approach to address these challenges by enhancing network correlations across global terrain features, such as structure, appearance, shape, and differentiation from background uniformity. However, VSA studies remain sparse in this context. Previous Earth-based aerial VSA have lacked a focus on geological terrain~\cite{vsa_aerial1}, while YOCOv1~\cite{yoco} is the only work targeting generalizable multi-class celestial terrain detection in UDA settings.

YOCOv1 addresses celestial appearance challenges through intra-feature clustering, correlating activations responsible for detections and reducing domain shift in their appearance. This approach contrasts with earlier methods that extract object features before clustering, neglecting background features. While effective in single-class feature-rich scenes like terrestrial urban driving, these earlier methods struggle with multi-class celestial surfaces (e.g., Mars), where features such as craters, dunes, and mountains lack sufficient activation discernibility. Despite its conceptual strengths, YOCOv1's practicality is limited by the YOLOv3 architecture, which cannot achieve real-time execution on spacecraft hardware~\cite{profiling}.

%% file: figs/arch.tex
\begin{figure*}
    \centering
    \includegraphics[width=0.8\textwidth]{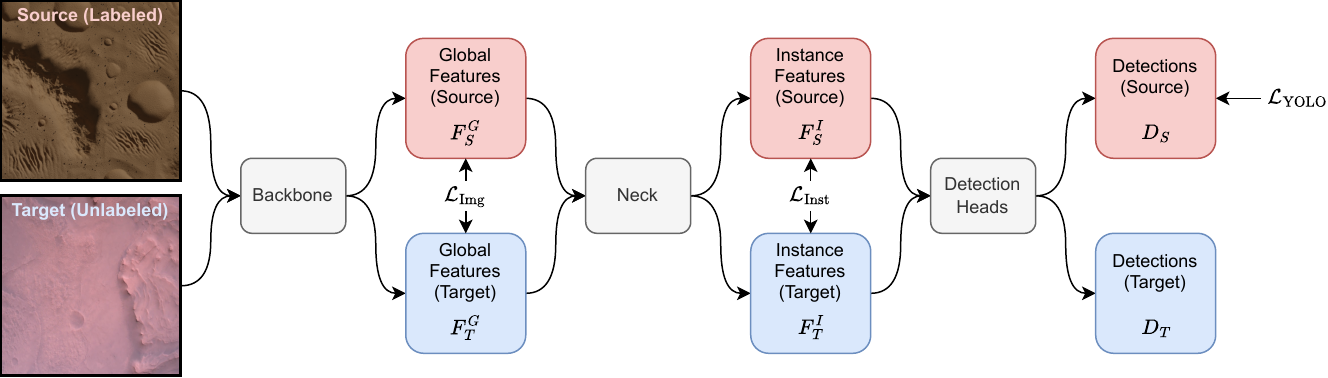}
    \caption{Overview of Unsupervised Domain Adaptation (UDA) within the YOLO object detection framework. Global image features and local instance features are aligned via $\img$ and $\inst$ respectively.}
    \label{fig:arch}
    \vspace{-1em}
\end{figure*}

%% file: sections/method.tex
\vspace{-1em}
\section{Methodology}\label{sec:method}
YOCOv1 marked a significant advancement in enhancing UDA performance for one-stage terrain detection but remains constrained by limitations that hinder its applicability to future missions. Building on the YOCOv1 framework, we propose an improved VSA strategy that accommodates a broader range of celestial environments and generalizes across YOLO architectures capable of real-time performance on current spacecraft hardware. In this section, we first describe the generalized YOLO architecture and modifications for UDA in~\autoref{sec:arch}, followed by our method for high-level global feature alignment in~\autoref{sec:global_align} and improved visual similarity-based instance alignment in~\autoref{sec:inst}.

\vspace{-1em}

\subsection{Network Architecture and UDA}\label{sec:arch}
The one-stage architecture used in YOCOv1, \textit{You Only Look Once} (YOLO), can be viewed generally as three distinct sub-networks, including a backbone feature extractor, neck, and three scale (small, medium, large) detection heads. YOLO has seen many new versions in recent years (e.g., v5~\cite{yolov5}, v6~\cite{yolov6}, v8~\cite{yolov8}) but has largely adhered to this paradigm. Our integration of UDA begins with input images consisting of labeled \textit{source} ($S$) and unlabeled \textit{target} ($T$) data, where source images constitute any arbitrary (although assumed to be style-similar) training dataset and target images are unlabeled real-world images from the desired operating domain. With features $F \in \mathbb{R}^{C \times H \times W}$ extracted from the backbone, we denote features from source and target imagery as $F_S$ and $F_T$ respectively and consider these features \textit{global} ($F_*^G$) as the backbone is a coarse processor. These global features are processed through the neck to produce local features $F_*^I$ which in turn seed the regression of bounding box, classification, and confidence properties to yield source and target detection matrices $D_S$ and $D_T$. The objective of UDA is to minimize the discrepancy between source and target features both globally and locally, enabling invariance to the domain and yielding accurate target detections $D_T$ without any explicit target supervision. Our learning objective for this alignment at the global (image) and local (instance) levels is denoted by $\img(F_S^G, F_T^G)$ and $\inst(F_S^I, F_T^I)$ respectively. The full learning objective of YOLO with image/instance UDA components is thus given by:
\begin{equation}
    \mathcal{L} = \mathcal{L}_{\text{YOLO}}(D_s) + \img(F^{G}_{S}, F^{G}_{T}) + \inst(F^{I}_S, F^{I}_{T})
\end{equation}
where $\mathcal{L}_{\text{YOLO}}$ is the traditional YOLO supervised loss. We refer readers to the YOLO literature (\cite{yolov5,yolov6,yolov8}) for the full definition. The architecture with UDA components is visualized in~\autoref{fig:arch}.

\vspace{-1em}

\subsection{Global Alignment ($\img$)}\label{sec:global_align}
UDA at the global feature level is facilitated by an adversarial training regime. Adversarial training learns a discriminator that predicts the originating domain of the features, where the goal of the network is to fool this discriminator into making the wrong predictions and induce feature-domain ambiguity. Specifically, global features from source $F^G_S$ and target $F^G_T$ domains are input into a discriminator $D$ which learns to predict the originating domain $d$ using a standard cross-entropy loss:
\begin{equation}\label{eq:adv}
    \mathcal{L}_{\text{Adv}}(F_d) = -d\log(D(F_d)) - (1-d)\log(1-D(F_d))
\end{equation}
where $d=0$ for source ($F^G_S$) and $d=1$ for target ($F^G_T$) data. To ensure the features from both domains become indistinguishable to the discriminator, the loss in~\autoref{eq:adv} is maximized with respect to $F_d$. This is achieved by using a Gradient Reverse Layer (GRL)~\cite{grl}, which applies a negative scaling factor to the gradients during backpropagation and reverses gradient flow. The GRL ensures that the discriminator minimizes the cross-entropy loss for domain classification, while the feature extractor maximizes it. The global image alignment loss $\img$ is given by:
\begin{equation}\label{eq:img}
    \img(F^G_S, F^G_T) = \adv(F^G_S) + \adv(F^G_T).
\end{equation}

\vspace{-1em}

\subsection{Visual Similarity-based Local Alignment ($\inst$)}\label{sec:inst}
The alignment of detection instance features is the main driver of UDA performance in object detection. To facilitate this in the context of challenging celestial surface imagery, we first propose a general-purpose regularization scheme that constrains domain drift on pre-regression features, a process which we refer to as Perceptual Consistency (PC,~\autoref{sec:pc}). Then, we examine two methodologies for minimizing object domain gap via VSA: \textit{Instance Clustering} (\autoref{sec:inst_clust}) and \textit{Feature Clustering} (\autoref{sec:feat_clust}).

\subsubsection{Perceptual Consistency Regularization}~\label{sec:pc}
% Previous literature has studied multi-scale discriminators~\cite{multi_scale1, multi_scale2} as a means of enhancing the adversarial training process by incorporating a feature-matching component.
Previous studies~\cite{multi_scale1, multi_scale2} have explored multi-scale discriminators to enhance adversarial training with feature matching. This approach extracts features from layers of the discriminator over each domain and aligns these features by minimizing the mean absolute error, which stabilizes the training process and improves performance by accounting for fine-grained details. As YOLO outputs (pre-regression) detection features at multiple scales directly, a similar regularization can be applied. 
We denote this regularization as Perceptual Consistency (PC) and define a $\pc$ loss as:
\begin{equation}
    \pc(F_S^I, F_T^I) = \sum_{i=1}^{|K|} \frac{1}{w_i} || F_{S_i}^{I} - F_{T_i}^{I} ||_1
\end{equation}
where the set of scales $K = (\text{large, medium, small})$, $|K| = 3$ and $F_{*_i}^{I}$ represent the features at scale $i$. Weights $w_i = 2^{|K|-i}$ rank the importance of each scale with an emphasis on smaller spatial resolutions.

\input{figs/inst}
\subsubsection{Instance Clustering VSA}\label{sec:inst_clust}
\textit{Instance Clustering} VSA takes as input the instance-level features for both source ($F^I_S$) and target ($F^I_T$) domains along with regressed detections $D_S$ and $D_T$. We adopt the ViSGA formulation~\cite{visga} to the one-stage setup, beginning by extracting instance-only features via the bounding box parameters, where each instance is flattened to yield fixed instance-feature embeddings $z \in \mathbb{R}^{N \times m}$. Hierarchical clustering is then employed over the set of source and target feature embeddings to create visually similar groupings; a flexible approach that dynamically assigns cluster centers instead of fixing a pre-determined amount as in K-Means. The average of each cluster is taken to construct a group-wise representative, which is input to instance discriminators and adversarially learned in a similar fashion to~\autoref{eq:adv}.

An alternative to adversarial training is the application of max-margin contrastive losses found common in the metric learning literature. Here, we again follow the ViSGA~\cite{visga} formulation and match embeddings from one domain to the nearest-neighbor ($nn$) embedding from the other domain and minimize the distance between their representations:
\vspace{-1em}
\begin{equation}
    \text{nn}(i) = \text{argmin}_{j < N_T} ||z_{S}^{i} - z_T^{j}||
\end{equation}
\begin{align}
\begin{split}
    \mathcal{L}_\text{Con} &= \sum_{i}^{|Z_S|} \left[ ||z_{S}^{i} - z_{T}^{\text{nn}(i)}||_2^2 \right. \\ 
    &+  \sum_{j, j \neq \text{nn}(i)}^{|Z_T|} \left.max\{ 0, m-||z_{S}^{i} - z_{T}^{j}||_2^2 \}\right]
\end{split}
\end{align}
where $Z_S$ and $Z_T$ are the set of source and target embeddings.  

The ability of hierarchical clustering to facilitate VSA is entirely dependent on the distinguishability of the instance embeddings. Poor disparity in the embedding space leads to loosely coupled groupings and unclear decision boundaries. This becomes incredibly prevalent in the context of celestial surface imagery, where we assume an already less distinctive feature space. To bolster instance distinction and downstream clustering performance, we propose to filter away features that may lead to ambiguous inter-similarities, and ensure that only \textit{strong}, robust features that truly represent the terrain of interest persist into the embeddings. Our technique for Strong Feature Filtering (SFF) begins by ranking each channel via channel-wise attention~\cite{ptap} of the instance-level feature $F^I_*$ to obtain feature weights $F_w$:
\begin{equation}\label{eq:sff}
    F_w = \sigma(\text{Conv1D}(\text{GAP}(F^I_*)))    
\end{equation}
where Conv1D is a one-dimensional convolution and GAP is the Global Average Pooling operator~\cite{gap}. The channel rankings $F_w$ are then applied to $F^I_*$ and sorted to yield ranked features $F_w'$, where we select the Top-K highest ranks to obtain filtered features $\hat{F}^I_*$. 
% An overview of the SFF module is shown in~\autoref{fig:sff}, while the complete pipeline for \textit{Instance Clustering} VSA is shown in~\autoref{fig:clust_inst}.

\subsubsection{Feature Clustering VSA}\label{sec:feat_clust}
As an alternative to extracting instance embeddings, YOCOv1 claims that the activations present within the channels of $F^I_*$ are a powerful identifier of detection class, and hypothesizes that forming clusters channel-wise will exhibit the same terrain distinctions with added residual background information that can assist alignment. This is largely inspired by prior work in contrastive learning dissection~\cite{yoco_insp}, where groupings of intermediary CNN layers were shown to form clear object boundaries. However, this approach proved sensitive to multi-class data, where intra-feature intricacies and background residuals were often indistinguishable. We hypothesize that intra-feature clustering could still be effective for certain celestial scenes where class-to-background distinctions are more prominent. Therefore, we explore two additional intra-feature clustering techniques: traditional K-Means clustering and a pixel pooling method based on self-attention similar to SFF.

% This proved to be incredibly sensitive to multi-class data, where intra-feature intricacies between multiple classes and background residuals were non-discernible and overly ambiguous. We hypothesize however that intra-feature clustering may still hold merit on celestial scenes where class-to-background distinctions are prominent. For this reason, we further examine two additional techniques at intra-feature clustering that propose to limit the amount of ambiguity exhibited by YOCOv1: traditional K-Means clustering and a pixel pooling method based on similar self-attention to SFF.

YOCOv1 uses hierarchical clustering with a preset number of groups, $C+1$, where $C$ is the number of object classes and one additional group is used for background. Alternatively, K-Means can be applied with $K=2$, distinguishing object activations from background activations. After clustering (either hierarchical or K-Means), channel-wise groupings are averaged and input to adversarial discriminators. Additionally, we explore the use of Pixel-wise Top-K Attention Pooling (PTAP~\cite{ptap}), which reduces features $F \in \mathbb{R}^{C \times H \times W}$ to $F \in \mathbb{R}^{1 \times H \times W}$, retaining the most important contributing pixels. Originally designed for co-localization, we adapt PTAP for a process similar to SFF, first applying channel attention to assign importance weights to each channel of $F_*^I$. The weighted feature map $F_w$ is then broadcast spatially and the sorted rankings are top-k selected \textit{pixel-wise}:
\begin{equation}\label{eq:spp}
    \hat{F}^I_*(x,y) = \frac{1}{|K|} * \sum_{j \in K}F_w \odot F^I_*(x,y,j)
\end{equation}
where $K$ is the number of channels to consider in the pooling operation and $x,y$ are each pixel location of the feature map. 
Each proposed scheme for VSA including PC and SFF components can be seen in~\autoref{fig:inst}.

% An overview of the techniques used for \textit{Feature Clustering} are given in~\autoref{fig:clust_feat}.

% YOCOv1 leverages the same hierarchical clustering procedure outline in~\autoref{sec:inst_clust}, although with a preset number of groups corresponding to $C+1$, or the number of classes plus an additional background catch-all class. Instead, K-Means can be leveraged with a generalization on all object classes, with the intuition that all object activations within a feature will be distinctive against background activations, such that the number of cluster centers is simply two (i.e., $K=2$). Post clustering (either by hierarchical or K-Means) results in channel-wise groupings, which are subsequently averaged together and fed to adversarial discriminators. As an alternative, we further examine the use of Pixel-wise Top-K Attention Pooling (PTAP~\cite{ptat}) to go directly from features $F \in \mathbb{R}^{C \times H \times W}$ to $F \in \mathbb{R}^{1 \times H \times W}$ with the most important contributing pixels retained. Originally proposed for co-localization, we adapt PTAP to achieve a similar process to SFF. Like SFF, we first apply channel attention to assign importance weights to each channel of $F_*^I$. 

%% file: figs/inst.tex
\begin{figure*}
    \centering
    \includegraphics[width=0.8\textwidth]{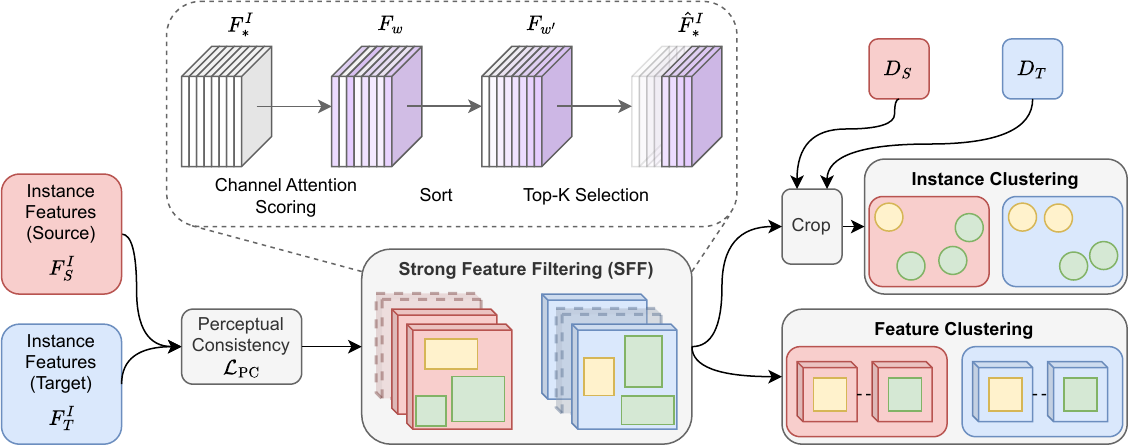}
    \caption{Proposed method for Visual Similarity-based Alignment (VSA) of object instances. \textit{Instance Clustering} minimizes the discrepancy between extracted source/target detections, while \textit{Feature Clustering} correlates individual feature maps by activation similarity.}
    \label{fig:inst}
    \vspace{-2em}
\end{figure*}

%% file: sections/eval.tex
\vspace{-1em}

\section{Evaluation}
Our evaluation begins with benchmarking YOLO version-size variants on spacecraft compute hardware to assess their feasibility for real-time, in-situ deployment, as detailed in~\autoref{sec:model_bench}. We then describe the datasets used for UDA evaluation, followed by implementation details and an analysis of the quantitative results in~\autoref{sec:quant}. Lastly, we conduct qualitative experiments using real-world data in~\autoref{sec:qual}.
% Our evaluation begins with a benchmark of YOLO version-size variants on spacecraft computer hardware in order to determine the feasibility of YOLO models for real-time and in-situ deployment, shown in~\autoref{sec:model_bench}. We then describe the datasets used for UDA evaluation, followed by implementation details and an analysis of the quantitative results in~\autoref{sec:quant}. Finally, we perform qualitative experiments on real-world data in~\autoref{sec:qual}.

% Next, we describe our implementation details for UDA including training setup and hyperparameter selection, and a description of the methods evaluated in~\autoref{sec:imp}. We then analyze the quantitative results in~\autoref{sec:quant}, and perform qualitative experiments on real-world data in.

% To evaluate our proposed formulation of VSA instance clustering and the performance of UDA on challenging celestial surface imagery, we start by

\vspace{-1em}

\subsection{Spacecraft Hardware Benchmarks}\label{sec:model_bench}
\input{tables/models}
The YOLO model family offers various version and size variants, each optimized for different hardware configurations and performance needs. 
% In this study, we evaluate the feasibility of YOLO inference on current spacecraft hardware to achieve real-time or near-real-time processing. 
We profile many of these variants on a flight hardware analog, the TUL PYNQ-Z2 evaluation board, which features a Zynq-7020 SoC with a dual-core ARM Cortex-A9 processor and an Edge TPU USB Accelerator connected via USB 2.0. This setup mirrors the NASA SpaceCube Mini-Z~\cite{sc}. ~\autoref{tab:models} presents YOLO models capable of real-time operation on this platform, with performance metrics including model disk size (MB), number of parameters (M), floating point operations (FLOPs, G), and inference latency (ms) for both CPU and Zynq hardware.

% The YOLO model family includes a variety of version and size variants, each tailored for different hardware configurations and performance requirements. In this study, we examine the feasibility of YOLO inference on current spacecraft hardware to achieve real-time or near-real-time processing rates. We perform profiling on a flight hardware analog, the TUL PYNQ-Z2 evaluation board, which is equipped with a Zynq-7020 SoC featuring a dual-core ARM Cortex-A9 processor and an Edge TPU USB Accelerator connected via USB 2.0. This setup closely mirrors the hardware architecture of the NASA SpaceCube Mini-Z~\cite{sc}.~\autoref{tab:models} presents the YOLO models that are capable of running in real-time on this platform. For each model, we report performance metrics including model disk size (MB), number of parameters (M), floating point operations (FLOPs, G), and inference latency (ms) on both the CPU and Zynq hardware. 

Of particular interest is the observation that only the nano (N), small (S), and medium (M) size models of YOLO v5, v6 (excluding medium), and v8 can achieve inference times under one second. Previous research~\cite{profiling} indicates that this significant drop in performance compared to desktop CPU architectures is primarily due to context switching and memory transfer bottlenecks between the co-processor (TPU) and ARM CPU. Large models, with substantial disk size (MB), require parameter streaming to the TPU for execution, which severely impacts inference time. However, we conclude that the majority of the evaluated variants are real-time capable and use these models as baselines for our UDA evaluation.
% Of particular interest, we see that only the nano (N), small (S), and medium (M) size models of YOLO v5, v6 (excluding medium), and v8 can execute an inference in under one second. Previous research~\cite{profiling} shows that this large decrease in execution performance compared with CPU architectures is due to the context switching and memory transfer bottleneck between the co-processor (TPU) and the ARM CPU, where large models in disk size (MB) have their parameters streamed to the TPU for execution, severely degrading inference time. However, we can conclude that the large majority of variants are real-time capable, and we leverage these models as the baselines for our evaluation of UDA.

% \subsection{Implementation Details}\label{sec:imp}

\vspace{-1em}

\subsection{Quantitative Evaluation}\label{sec:quant}
 \input{tables/map_1/map_mars}
 \input{tables/map_1/map_asteroid}
For quantitative evaluation, we utilize six distinct datasets across three representative space environments: Mars, Asteroid, and Moon, and report the Mean Average Precision (mAP) at an Intersection-over-Union (IoU) threshold of 0.5 (mAP@0.5).

\subsubsection{Datasets}

\input{figs/data.tex}
As no labeled instances of surface terrain imagery exist for Mars and Asteroid, we generate two distinct synthetic image datasets using the Blender 3D software with varying texture and material assets, referred to as \textit{Sim Mars} and \textit{Sim Asteroid}. \textit{Sim Mars} contains ground truth labels for the classes \textit{Crater}, \textit{Dune}, and \textit{Mountain}, while \textit{Sim Asteroid} is a single-class dataset with \textit{Boulders}. For the Moon, we use NASA mission-derived global mosaics of the lunar surface~\cite{mosaic100} at 474 and 100 meters-per-pixel (m/px) resolutions as source and target datasets, respectively, and produce image frames via random crops. We utilize the Robbins Moon Crater Database~\cite{robbins} to correlate known crater impact locations with each frame, generating ground truth bounding box labels. This setup reflects a realistic mission concept of operations, as the exploration of new bodies often involves gradually increasing data resolution. The 474 m/px (low resolution) data is used for training, while the 100 m/px (high resolution) data is used for testing. Each Moon dataset contains a single class, \textit{Crater}. Example training and testing frames for \textit{Moon}, \textit{Sim Mars}, and \textit{Sim Asteroid} are shown in~\autoref{fig:data}.
% As no labeled instances of surface terrain imagery exist for Mars and Asteroid, we develop two distinct sets of synthetic images utilizing the Blender 3D software with varying texture and material assets, which we denote as \textit{Sim Mars} and \textit{Sim Asteroid} respectively.~\textit{Sim Mars} contains ground truth labels for \textit{Crater}, \textit{Dune}, and \textit{Mountain} classes, while \textit{Sim Asteroid} is a single-class dataset consisting of \textit{Boulders}. In the case of the Moon, we take NASA mission-derived global mosaics of the lunar surface~\cite{mosaic100} at 474 and 100 meters-per-pixel (m/px) as the source/target datasets and produce image frames by random crops. We leverage the Robbins Moon Crater Database~\cite{robbins} to provide latitude, longitude, and diameters of known crater impacts and correlate these with each frame to yield ground truth bounding box labels. This experiment represents a realistic mission concept of operations, as the discovery and exploration of new bodies often involve gradually increasing data resolution. The 474 m/px (low resolution) data is used as training while the 100 m/px (high resolution) imagery is used for testing. Each Moon dataset has one class of \textit{Crater}. Example training and testing frames for \textit{Moon} as well as \textit{Sim Mars} and \textit{Sim Asteroid} are showcased in~\autoref{fig:data}.

 \subsubsection{Implementation Details}\label{sec:imp}
 We utilize the Ultralytics~\cite{ultralytics} open-source implementation of all YOLO variants and implement our UDA components as plugins for ease of use and extensibility to any model available on the platform. We train all models from scratch without incorporating any pre-trained weights. Training was conducted on a single NVIDIA A6000 GPU for 50 epochs with a batch size of 32. 

We study 11 loss variations across the models benchmarked in~\autoref{tab:models}. The baseline configurations are \textit{Source Only} and \textit{Target Only}, representing YOLO models trained exclusively on source or labeled target data, respectively. The \textit{Target Only} model serves as an oracle, indicating performance without any data distribution gap. Nine additional models perform UDA, categorized based on the VSA types described in~\autoref{sec:method}: adversarial instance clustering (\textit{Instance, Adversarial}), contrastive instance clustering (\textit{Instance, Contrastive}), and adversarial feature clustering (\textit{Feature, Adversarial}). For \textit{Instance} methods, we examine three loss configurations: (i) the original ViSGA~\cite{visga} hierarchical clustering procedure adapted for YOLO (\textit{ViSGA}), (ii) hierarchical clustering with PC loss (\textit{PC Only}), and (iii) a combination of hierarchical clustering, PC, and SFF modules (\textit{PC SFF}). For \textit{Feature} methods, we explore three loss configurations: (i) YOCOv1's intra-feature hierarchical clustering~\cite{yoco} (\textit{YOCOv1}), (ii) K-Means clustering with PC loss (\textit{PC K-Means}), and (iii) PTAP's pixel pooling approach~\cite{ptap} (\textit{PTAP}). For all \textit{Instance}-based losses, we utilize agglomerative clustering with the maximum linkage, cosine as the distance function, and a cluster merging threshold of 0.1. 
% For \textit{YOCOv1}, agglomerative clustering with $C+1$ groups (number of classes plus a ``background'') and the average linkage is used. For \textit{PC K-Means}, we define $K=2$ for ``object'' and ``background'' groupings. 
For \textit{PC SFF} and \textit{PTAP}, we set the feature selection threshold to the top 50\%.

\subsubsection{Sim Mars}
Accuracies for all model variants across the synthetic Mars scenes are shown in~\autoref{tab:map_mars}, with the best-performing method for each architecture-size variant highlighted in green. Of first note, \textit{Instance} clustering methods consistently outperform \textit{Feature} clustering methods in most models, except for the YOLO 6-N variant where \textit{YOCOv1} achieves the highest performance. Within the \textit{Instance} category, contrastive learning (\textit{Instance, Contrastive}) is the most effective approach, achieving the highest accuracy in four out of eight models, while adversarial training (\textit{Instance, Adversarial}) performs best in three models. This suggests that contrastive learning may better handle multi-class alignment, as discriminator networks struggle to accurately distinguish instances in the case of the Mars surface.

Additionally, a comparison of the adversarial and contrastive instance methods reveals that PC SFF outperforms ViSGA in most cases. Of the seven top-performing models in this category, PC SFF exceeds ViSGA in five. ViSGA only performs better with YOLO 5-M and 8-M. Notably, PC SFF shows strong generalizability, particularly in low-parameter models like YOLO 5-S, 6-S, and 8-S, with accuracies of 66.3\%, 69.6\%, and 66.6\%, respectively.

Compared to \textit{Source Only} training, PC SFF provides significant performance improvements, particularly for YOLO 5-N, where it achieves an 11\% accuracy increase, highlighting its effectiveness in making low-performing models viable through VSA with robust features.

\vspace{-1em}
\subsubsection{Sim Asteroid}
\input{tables/map_1/map_moon}
Similar results are observed in~\autoref{tab:map_asteroid} for \textit{Sim Asteroid}, where the proposed PC SFF technique outperforms the traditional ViSGA approach in instance clustering methods, surpassing it in four of eight cases. Notably, Perceptual Consistency alone performs best in two YOLO variants (5-N and 6-S), suggesting a potential detrimental learning effect in PC SFF's channel attention scoring mechanisms, particularly in YOLO 5-N, where performance drops by approximately 38\% between \textit{PC Only} and \textit{PC SFF}. In contrast to \textit{Sim Mars}, Feature clustering methods excel in three models, indicating their potential utility in single-class scenarios. ViSGA does not achieve the highest accuracy for any YOLO model, while YOCOv1 and adversarial PC SFF tie on YOLO 8-S. Among feature methods, K-Means with PC performs best in two instances, achieving the highest accuracy of 67\% with the YOLO 6-N model. This suggests the effectiveness of intra-feature clustering in single-class scenes, where a $K=2$ division of objects and backgrounds is highly effective.

The low performance of certain \textit{Source Only} YOLO models underscores the challenging nature of the Asteroid dataset. For example, YOLO 5-N achieves only 5.4\%, while YOLO 8-M and 8-N achieve just 25\%. This highlights the critical importance of VSA techniques in addressing the highly similar and texture-repetitive feature space in UDA. Compared to \textit{Source Only} training, \textit{PC Only} results in a substantial accuracy improvement of about 48\% for YOLO 5-N in the adversarial setting, with \textit{PTAP} emerging as the next best-performing approach. This validates the effectiveness of pixel pooling for low-parameter models, where per-pixel object distinction is easier. Furthermore, significant improvements are observed in instances where \textit{ViSGA} underperforms, such as with YOLO 8-S and \textit{Instance, Contrastive}, where \textit{ViSGA} achieves 22.1\% accuracy and PC SFF achieves 53\%, indicating an improvement of nearly 31\%

\vspace{-1em}
\subsubsection{Moon}
\input{figs/qual/moon_gt}
\input{figs/qual/mars}
Results for the low-to-high resolution real-world lunar data adaptation are shown in~\autoref{tab:map_moon}. Firstly, low accuracy of the \textit{Target Only} (oracle) models is observed, indicating erroneous or malformed ground truth. This can be attributed to the Robbins lunar crater database being incomplete, as it does not provide ground truth for every crater on the lunar surface. This discrepancy is evident in~\autoref{fig:qual_moon_gt}, where predictions from the baseline YOLO 8-M appear more complete than the provided ground truth. Consequently, the mAP calculation underestimates the true detection capability of the models.

Despite this limitation, a slight increase in performance is observed across all models when comparing \textit{Source Only} to \textit{Target Only} training, suggesting that the difference in resolution between source and target data constitutes a domain gap and justifies the necessity for UDA. Among these, the \textit{Instance, Contrastive} category achieves the highest performance in six out of eight models. This result contrasts the \textit{Sim Asteroid} experiment which is similarly single-class, where contrastive learning performed poorly. This demonstrates a general measure of the task difficulty, as we can assume it is easier to distinguish resolution differences within an embedding space as opposed to adversarial discriminators. Within the \textit{Instance, Contrastive} methods, the original ViSGA implementation achieves the highest performance in only one instance, with the remaining improvements driven by \textit{PC Only} (YOLO 6-S, 8-N) and \textit{PC SFF} (YOLO 5-S, 5-M, 6-N). An additional noteworthy observation is that \textit{Source Only} training retains the best performance on YOLO 5-N, a potential limitation in representation and learning capacity that prevents domain adaptation from discovering an effective alignment scheme.

% This highlights the advantage of scoring and selecting features based on learnable channel attention, which enables the extraction of robust representative features that enhance downstream detection and domain alignment tasks. 

\vspace{-1em}
\subsection{Qualitative Results}\label{sec:qual}
In this section, we evaluate the qualitative performance on real-world data. 
% We first begin by discussing the datasets and model variants used in this experiment, followed by a visualization of the results. 
Our real-world dataset for Mars is Mars HiRISE Landmarks~\cite{hirise}, which showcases cropped instances of terrain features on the Martian surface. Our Asteroid data is prior mission imaging from OSIRIS-REx of the Asteroid 101955 Bennu~\cite{orex_data}. For Moon, we utilize the same 100 m/px Moon dataset as in the quantitative experiments as this data is already real-world. We showcase YOLO 5-M \textit{Instance, Adversarial} methods including \textit{ViSGA}, \textit{PC Only}, and \textit{PC SFF} for Mars, the same methods although \textit{Instance, Contrastive} and YOLO 8-M on Moon, and YOLO 6-S \textit{Feature, Adverarial} methods including \textit{YOCOv1}, \textit{PC K-Means}, and \textit{PTAP} on Asteroid. We train each setup with the same source datasets used in the quantitative experiments. All qualitative results have inference detection parameters set at $\geq 0.25$ confidence and an IoU threshold of 0.7.

\autoref{fig:qual_mars} showcases the qualitative results on Mars, with red representing the \textit{Crater} class, blue representing the \textit{Mountain} class, and green representing the \textit{Dune} class. Training with \textit{Source Only} data leads to numerous incorrect detections, such as false crater detections in the top row, mountain detections in blackened areas in the middle row, and misidentified sand and mountain features in the bottom row. ViSGA provides only marginal improvements, still yielding many erroneous detections. 
% Notable improvements are seen in the third image (bottom row), where erroneous sand detections in the \textit{Source Only} model are partially corrected to individual crater detections.
\textit{PC Only} demonstrates enhanced performance over ViSGA, with improvements particularly evident in the first and second images (top and middle rows). \textit{PC SFF} achieves the best qualitative results on this dataset, significantly reducing erroneous detections present in both the \textit{Source Only} and ViSGA models. This configuration yields accurate new detections while simultaneously minimizing erroneous ones, as demonstrated in all three images.
% \autoref{fig:qual_mars} showcases the qualitative results on Mars. Here, red represents the \textit{Crater} class, blue represents the \textit{Mountain} class, and green represents the \textit{Dune} class. Here we observe that training with \textit{Source Only} data leads to numerous incorrect detections, as illustrated by examples such as false crater detections in the top row, mountain detections in blackened areas in the middle row, and misidentified sand and mountain features in the bottom row. ViSGA provides only a marginal improvement that still yields many erroneous detections. Notable improvements can be observed in the third image (bottom row), where erroneous sand detections from the \textit{Source Only} model are partially corrected to individual crater detections.~\textit{PC Only} demonstrates enhanced performance over ViSGA, where improvements are particularly evident in the first and second images (top and middle rows, respectively).~\textit{PC SFF} achieves the highest qualitative results on this dataset. This configuration significantly reduces erroneous detections present in the \textit{Source Only} and ViSGA models, yielding accurate \textit{new} detections while simultaneously giving \textit{less} erroneous detections, evident in all three examples.

\input{figs/qual/asteroid}
\input{figs/qual/moon}
\autoref{fig:qual_asteroid} demonstrates qualitative results on the OSIRIS-REx imagery where red detections represent the single class \textit{Boulder}. Similar to the results observed on the Mars dataset, \textit{Source Only} training produces numerous erroneous and incorrect detections.~\textit{YOCOv1} demonstrates improved detections in this challenging scene, as shown most decisively in the third example (bottom row). The \textit{PC K-Means} method achieves relatively comparable performance, although issues such as overlapping and multiple detections of the same object are apparent in the first example where absent from \textit{YOCOv1}. Additionally, in the third example, many subjectively correct detections are removed, highlighting the limitations of two-class clustering. The learnable pooling of \textit{PTAP} strikes a balance between these extremes. Erroneous detections are significantly reduced while retaining high-quality detections, yielding more reliable and consistent results.

\autoref{fig:qual_moon} portrays qualitative results for the Moon. In this case, we observe significantly stronger detection capabilities across all models, including \textit{Source Only}. This improvement can be attributed to the less challenging gap in data distribution.
% as the differences in resolution contrast with the more complex texture and color spaces of previous datasets. 
The results mirror those from previous experiments, where \textit{Source Only} exhibits some erroneous and overlapping detections. The \textit{ViSGA} method provides a slight improvement, followed by further enhancements with the \textit{PC Only} and \textit{PC SFF}.~\textit{PC SFF} achieves the best subjective performance, demonstrating a marked reduction in erroneous detections while retaining high-quality, accurate detections. This method effectively eliminates many of the errors observed in both \textit{Source Only} and \textit{ViSGA} outputs. Overall, PC SFF shows very convincing results, detecting a roughly complete set of crater detections.

% For Asteroid and Moon, red represents the single classes of \textit{Boulder} and \textit{Crater} respectively. 

% \autoref{fig:qual_mars} demonstrates qualitative results across three HiRISE dataset images for the \textit{Instance, Adversarial} methods. For these visualizations,  \red{conf iou setting}. 

% \autoref{fig:qual_asteroid} demonstrates qualitative results across three OSIRIS-REx Bennu images demonstrating Full Feature methods of \textit{Source Only}, \textit{YOCOv1}, \textit{PC K-Means}, and \textit{PTAP} approaches on the YOLO 6-S model. Here, \textit{red} bounding boxes represent detections of the \textit{boulder} class.  \red{conf iou setting}. Similar to the results observed on the Mars dataset, \textit{Source Only} training produces numerous erroneous and incorrect detections. The \textit{YOCOv1} approach demonstrates improved detection of boulders in this challenging scene, as evident in the third example (bottom row).

%% file: tables/models.tex
\begin{table}[h!]
\centering
\vspace{-2em}
\caption{Comparison of YOLO model versions and architecture sizes considered in this work, including execution time on spacecraft (Zynq) and CPU hardware.
}
\resizebox{\linewidth}{!}{%
\begin{tblr}{
  width = \linewidth,
  colspec = {Q[c,m,120]Q[c,m,120]Q[c,m,120]Q[c,m,120]Q[c,m,200]Q[c,m,200]},
  hline{1,2,10} = {-}{},
  hline{5,7} = {-}{dashed},
}
Model & Size (MB) & Params (M) & FLOPs (G) & {Latency\\(Zynq, ms)} & {Latency\\(CPU, ms)}\\
5-N & \phantom{0}3.2 & \phantom{0}2.7 & \phantom{0}1.2 & \phantom{0}79 & 22.6\\
5-S & \phantom{0}9.2 & \phantom{0}9.2 & \phantom{0}3.9 & 143 & 22.3\\
5-M & 25.1 & 25.1 & 10.3 & 736 & 22.7\\
6-N & \phantom{0}4.8 & \phantom{0}4.5 & 13.1 & \phantom{0}83 & 23.0\\
6-S & 16.6 & 16.5 & 44.9 & 428 & 23.1\\
8-N & \phantom{0}3.6 & \phantom{0}3.2 & \phantom{0}9.6 & \phantom{0}81 & 22.2\\
8-S & 11.1 & 11.2 & 31.6 & 205 & 22.3\\
8-M & 26.6 & 25.9 & 87.0 & 784 & 22.5
\end{tblr}
}
\label{tab:models}
\end{table}

%% file: tables/map_1/map_mars.tex
\begin{table*}
\centering
\caption{VSA accuracy (mAP@0.5) on \textit{Sim Mars}.}
\label{tab:map_mars}
\begin{tblr}{
  width = \linewidth,
  % Used to be 60s
  colspec = {Q[1]Q[c,m,20]Q[c,m,25]Q[c,m,35]Q[c,m,35]Q[c,m,35]Q[c,m,35]Q[c,m,35]Q[c,m,35]Q[c,m,35]Q[c,m,35]Q[c,m,35]Q[c,m,25]},
  cells = {c},
  cell{1}{3} = {c=11}{},
  cell{2}{3} = {r=2}{},
  cell{2}{4} = {c=3}{},
  cell{2}{7} = {c=3}{},
  cell{2}{10} = {c=3}{},
  cell{2}{13} = {r=2}{},
  cell{4}{1} = {r=8}{},
  vline{3} = {4-11}{},
  vline{4,7,10,13} = {2-11}{dashed},
  hline{1} = {-}{},
  hline{3} = {4-12}{dashed},
  hline{4} = {3-13}{},
  hline{12} = {-}{},
}
 &  & \textbf{Method} &  &  &  &  &  &  &  &  &  & \\
 &  & {Source\\Only} & Instance, Adversarial &  &  & Instance, Contrastive &  &  & Feature, Adversarial &  &  & {Target\\Only}\\
 &  &  & ViSGA \cite{visga} & PC Only & PC SFF & ViSGA \cite{visga} & PC Only & PC SFF & YOCOv1 \cite{yoco} & PC K-Means & PTAP \cite{ptap} & \\
\begin{sideways}\textbf{Architecture}\end{sideways}

& 5-N & 51.1 & 57.9 & 55.7 & \SetCell[c=1]{c, olive9}\textbf{62.4} & 56.4 & 53.6 & 52.2 & 59.3 & 54.3 & 58.4 & 93.7\\
& 5-S & 64.0 & 65.4 & 64.0 & 64.6 & 60.7 & 60.9 & \SetCell[c=1]{c, olive9}\textbf{66.3} & 62.7 & 55.5 & 63.2 & 94.2\\
& 5-M & 63.2 & 68.3 & 71.4 & 66.1 & \SetCell[c=1]{c, olive9}\textbf{71.6} & 65.9 & 68.4 & 68.8 & 65.4 & 62.9 & 94.9\\
& 6-N & 64.9 & 58.2 & 56.7 & 56.1 & 55.2 & 59.2 & 55.8 & \SetCell[c=1]{c, olive9}\textbf{65.0} & 60.4 & 50.8 & 93.9\\
& 6-S & 60.9 & 64.3 & 66.0 & 64.3 & 65.7 & 69.2 & \SetCell[c=1]{c, olive9}\textbf{69.6} & 65.5 & 66.2 & 64.2 & 94.3\\
& 8-N & 57.4 & 51.9 & 55.9 & 51.7 & 58.7 & 51.4 & \SetCell[c=1]{c, olive9}\textbf{60.4} & 56.2 & 55.0 & 56.9 & 94.2\\
& 8-S & 57.6 & 65.0 & 65.7 & \SetCell[c=1]{c, olive9}\textbf{66.6} & 62.6 & 63.7 & 63.6 & 56.1 & 65.0 & 62.7 & 94.7\\
& 8-M & 65.0 & \SetCell[c=1]{c, olive9}\textbf{68.9} & 67.2 & 65.5 & 61.6 & 66.4 & 56.9 & 66.7 & 63.4 & 67.7 & 95.0\\

\end{tblr}
\vspace{-1em}
\end{table*}

%% file: tables/map_1/map_asteroid.tex
\begin{table*}
\centering
\caption{VSA accuracy (mAP@0.5) on \textit{Sim Asteroid}.}
\label{tab:map_asteroid}
\begin{tblr}{
  width = \linewidth,
  colspec = {Q[1]Q[c,m,20]Q[c,m,25]Q[c,m,35]Q[c,m,35]Q[c,m,35]Q[c,m,35]Q[c,m,35]Q[c,m,35]Q[c,m,35]Q[c,m,35]Q[c,m,35]Q[c,m,25]},
  cells = {c},
  cell{1}{3} = {c=11}{},
  cell{2}{3} = {r=2}{},
  cell{2}{4} = {c=3}{},
  cell{2}{7} = {c=3}{},
  cell{2}{10} = {c=3}{},
  cell{2}{13} = {r=2}{},
  cell{4}{1} = {r=8}{},
  vline{3} = {4-11}{},
  vline{4,7,10,13} = {2-11}{dashed},
  hline{1} = {-}{},
  % hline{2} = {-}{},
  hline{3} = {4-12}{dashed},
  hline{4} = {3-13}{},
  hline{12} = {-}{},
}
 &  & \textbf{Method} &  &  &  &  &  &  &  &  &  & \\
 &  & {Source\\Only} & Instance, Adversarial &  &  & Instance, Contrastive &  &  & Feature, Adversarial &  &  & {Target\\Only}\\
 &  &  & ViSGA \cite{visga} & PC Only & PC SFF & ViSGA \cite{visga} & PC Only & PC SFF & YOCOv1 \cite{yoco} & PC K-Means & PTAP \cite{ptap} & \\
\begin{sideways}\textbf{Architecture}\end{sideways}

& 5-N & \phantom{0}5.4 & 10.9 & \SetCell[c=1]{c, olive9}\textbf{53.1} & 15.0 & 37.6 & 39.9 & 26.2 & 38.0 & 32.9 & 47.6 & 92.0\\
& 5-S & 35.5 & 56.5 & 44.7 & 34.9 & 37.2 & 52.8 & 31.8 & 50.2 & \SetCell[c=1]{c, olive9}\textbf{57.0} & 35.7 & 92.4\\
& 5-M & 32.8 & 51.2 & 61.0 & \phantom{0}4.6 & 44.2 & 40.4 & \SetCell[c=1]{c, olive9}\textbf{61.2} & 44.7 & 49.5 & 53.5 & 92.6\\
& 6-N & 66.3 & 58.8 & 65.5 & 60.1 & 58.8 & 63.2 & 61.8 & 67.3 & \SetCell[c=1]{c, olive9}\textbf{67.4} & 62.8 & 91.9\\
& 6-S & 64.3 & 58.9 & \SetCell[c=1]{c, olive9}\textbf{66.2} & 65.7 & 58.9 & 65.3 & 63.0 & 59.8 & 62.4 & 60.7 & 92.8\\
& 8-N & 26.0 & 31.1 & 43.4 & \SetCell[c=1]{c, olive9}\textbf{60.7} & 40.3 & 17.8 & 55.3 & 34.0 & 43.3 & 24.1 & 91.8\\
& 8-S & 55.3 & 36.8 & 55.2 & \SetCell[c=1]{c, olive9}\textbf{56.1} & 22.1 & 45.5 & 53.0 & \SetCell[c=1]{c, olive9}\textbf{56.1} & 13.2 & 18.7 & 92.5\\
& 8-M & 25.2 & 49.0 & 49.5 & \SetCell[c=1]{c, olive9}\textbf{53.8} & 41.9 & 45.8 & 41.2 & 52.3 & 32.6 & 40.0 & 93.2\\

\end{tblr}
\vspace{-1em}
\end{table*}

%% file: figs/data.tex
\begin{figure}
    \centering
    \setkeys{Gin}{width=\columnwidth}
    % \begin{subfigure}{\dimexpr0.33\columnwidth+10pt\relax}
    \begin{subfigure}{0.32\columnwidth}
        % \makebox[10pt]{\raisebox{30pt}{\rotatebox[origin=c]{90}{Train}}}%
        % \includegraphics[width=\dimexpr\linewidth-10pt\relax]{figs/data/mars1.png}\\[3pt]
        \includegraphics{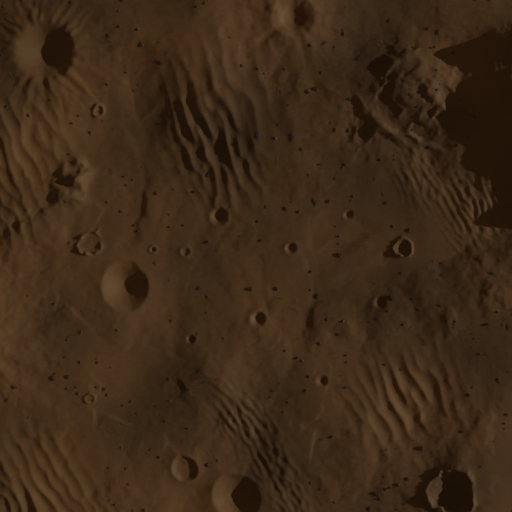}\\[3pt]
        % \makebox[10pt]{\raisebox{30pt}{\rotatebox[origin=c]{90}{Test}}}%
        % \includegraphics[width=\dimexpr\linewidth-10pt\relax]{figs/data/mars2.png}
        \includegraphics{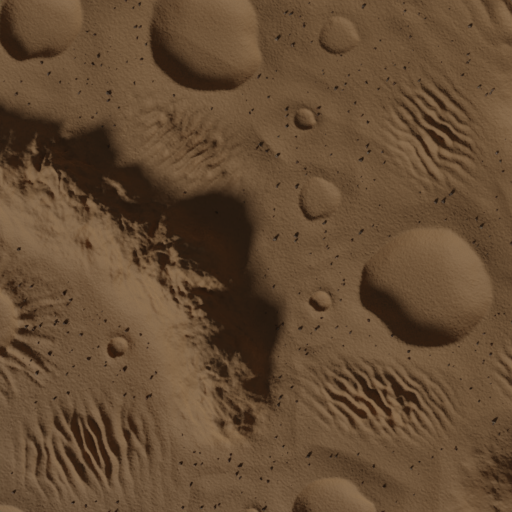}
        \caption{Mars}
    \end{subfigure}
    \hfil
    \begin{subfigure}{0.32\columnwidth}
        \includegraphics{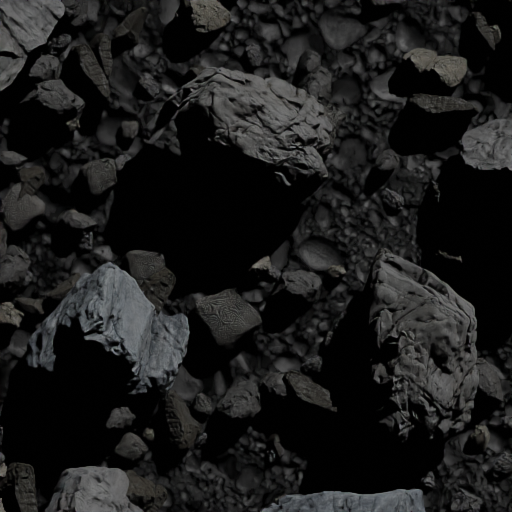}\\[3pt]
        \includegraphics{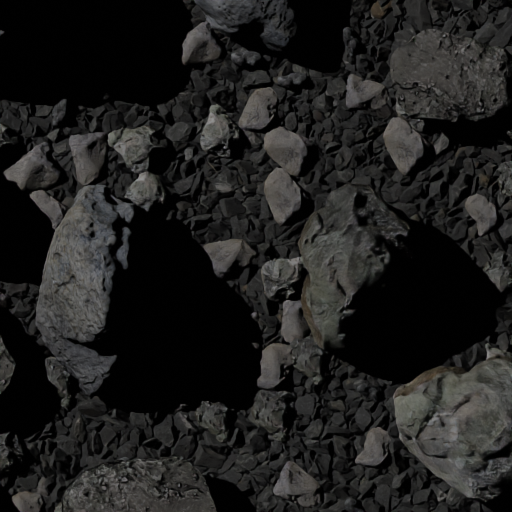}
        \caption{Asteroid}
    \end{subfigure}
    \hfil
    \begin{subfigure}{0.32\columnwidth}
        \includegraphics{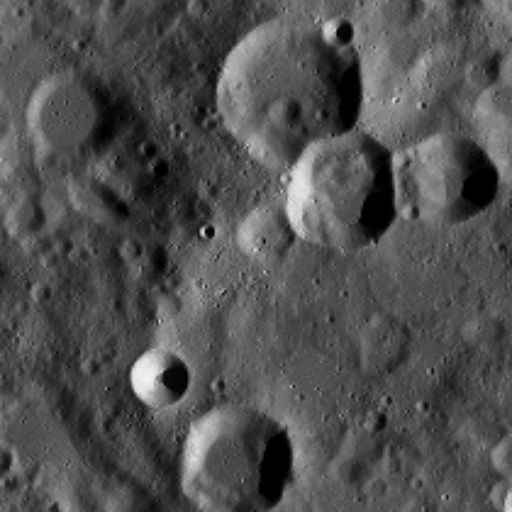}\\[3pt]
        \includegraphics{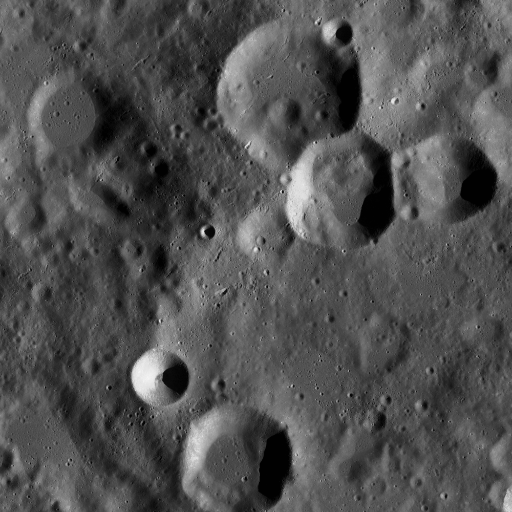}
        \caption{Moon}
    \end{subfigure}
    \caption{Mars, Asteroid, and Moon training (top row) and testing (bottom row) dataset examples.}
    \label{fig:data}
    \vspace{-2em}
\end{figure}

%% file: tables/map_1/map_moon.tex
\begin{table*}
\centering
\caption{VSA accuracy (mAP@0.5) on \textit{Moon}.}
\label{tab:map_moon}
\begin{tblr}{
  width = \linewidth,
  colspec = {Q[1]Q[c,m,20]Q[c,m,25]Q[c,m,35]Q[c,m,35]Q[c,m,35]Q[c,m,35]Q[c,m,35]Q[c,m,35]Q[c,m,35]Q[c,m,35]Q[c,m,35]Q[c,m,25]},
  cells = {c},
  cell{1}{3} = {c=11}{},
  cell{2}{3} = {r=2}{},
  cell{2}{4} = {c=3}{},
  cell{2}{7} = {c=3}{},
  cell{2}{10} = {c=3}{},
  cell{2}{13} = {r=2}{},
  cell{4}{1} = {r=8}{},
  vline{3} = {4-11}{},
  vline{4,7,10,13} = {2-11}{dashed},
  hline{1} = {-}{},
  hline{3} = {4-12}{dashed},
  hline{4} = {3-13}{},
  hline{12} = {-}{},
}
 &  & \textbf{Method} &  &  &  &  &  &  &  &  &  & \\
 &  & {Source\\Only} & Instance, Adversarial &  &  & Instance, Contrastive &  &  & Feature, Adversarial &  &  & {Target\\Only}\\
 &  &  & ViSGA \cite{visga} & PC Only & PC SFF & ViSGA \cite{visga} & PC Only & PC SFF & YOCOv1 \cite{yoco} & PC K-Means & PTAP \cite{ptap} & \\
\begin{sideways}\textbf{Architecture}\end{sideways}

& 5-N & \SetCell[c=1]{c, olive9}\textbf{34.7} & 34.2 & 34.0 & 34.6 & 33.9 & 33.1 & 33.7 & 33.5 & 34.6 & 34.6 & 36.9\\
& 5-S & 33.9 & 34.6 & 34.9 & 33.4 & 34.6 & 33.8 & \SetCell[c=1]{c, olive9}\textbf{35.2} & 34.5 & 34.6 & 34.6 & 37.5\\
& 5-M & 34.7 & 34.0 & 33.9 & 34.0 & 35.1 & 34.1 & \SetCell[c=1]{c, olive9}\textbf{35.6} & 33.4 & 34.8 & 35.0 & 37.4\\
& 6-N & 34.0 & 33.4 & 33.7 & 34.4 & 33.3 & 34.4 & \SetCell[c=1]{c, olive9}\textbf{35.2} & 34.0 & 34.0 & 33.5 & 35.6\\
& 6-S & 33.6 & 34.5 & 35.1 & 34.8 & 34.1 & \SetCell[c=1]{c, olive9}\textbf{35.2} & 33.7 & 34.1 & 34.7 & 34.4 & 36.6\\
& 8-N & 33.4 & 33.7 & 34.2 & 33.7 & 33.5 & \SetCell[c=1]{c, olive9}\textbf{35.2} & 34.3 & 33.5 & 33.0 & 33.0 & 36.4\\
& 8-S & 33.0 & 33.6 & 33.6 & 32.9 & 32.3 & 33.8 & 33.3 & 33.9 & \SetCell[c=1]{c, olive9}\textbf{34.7} & 34.1 & 36.9\\
& 8-M & 32.8 & 33.2 & 33.9 & 33.1 & \SetCell[c=1]{c, olive9}\textbf{35.0} & 34.0 & 33.8 & 33.6 & 32.6 & 33.1 & 37.1\\

\end{tblr}
\vspace{-1em}
\end{table*}

%% file: figs/qual/moon_gt.tex
% \begin{figure}
%     \centering
%     \subfigure[Ground Truth]{\includegraphics[width=0.49\columnwidth]{figs/qual/moon/moon_367_gt.png}}\hfill
%     \subfigure[YOLOv8-M Predictions]{\includegraphics[width=0.49\columnwidth]{figs/qual/moon/moon_367_stock.png}}
%     \caption{An example of model detections against known crater labels, demonstrating incomplete ground truth that artificially degrades the quantitative performance.}
%     \label{fig:qual-moon-gt}
% \end{figure}

\begin{figure}[!h]
    \centering
    \setkeys{Gin}{width=\columnwidth}
    \begin{subfigure}{0.49\columnwidth}
        \includegraphics{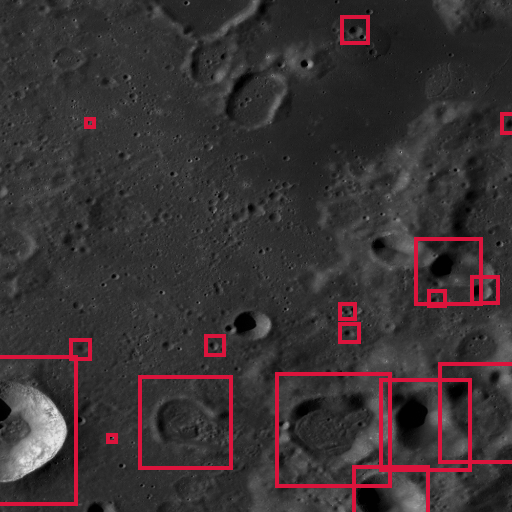}
        \caption{Ground Truth}
    \end{subfigure}
    \hfil
    \begin{subfigure}{0.49\columnwidth}
        \includegraphics{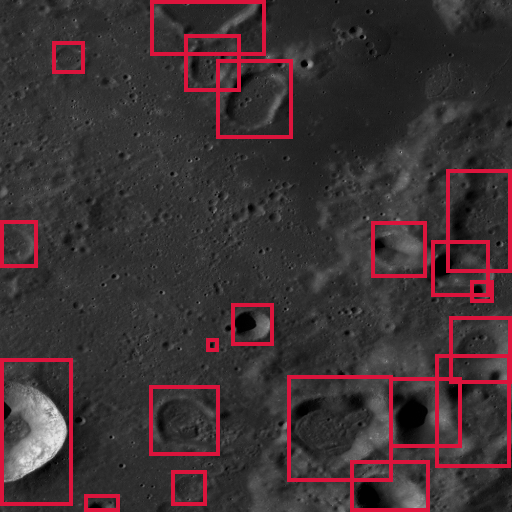}
        \caption{YOLOv8-M Predictions}
    \end{subfigure}
    \caption{Example of model detections against known crater labels, demonstrating incomplete ground truth that artificially degrades the quantitative performance.}
    \label{fig:qual_moon_gt}
    \vspace{-3em}
\end{figure}

%% file: figs/qual/mars.tex
\begin{figure*}
    \centering
    \setkeys{Gin}{width=\linewidth}
    \begin{subfigure}{\qualsize\textwidth}
        \includegraphics{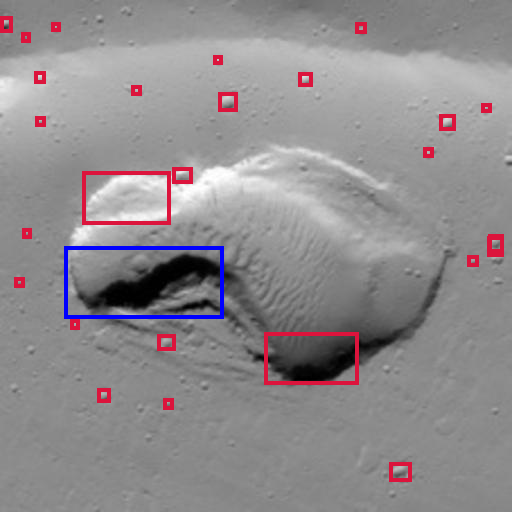}\\[3pt]
        \includegraphics{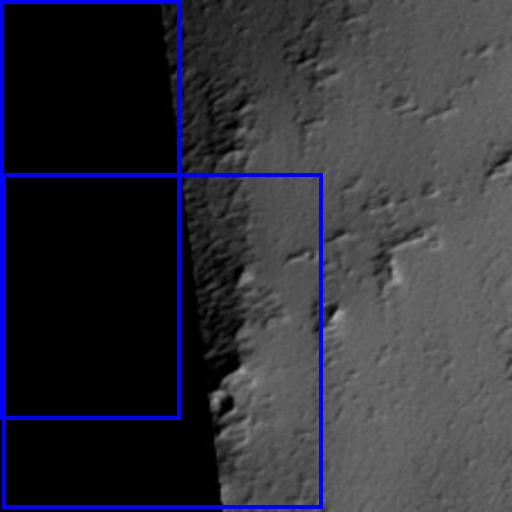}\\[3pt]
        \includegraphics{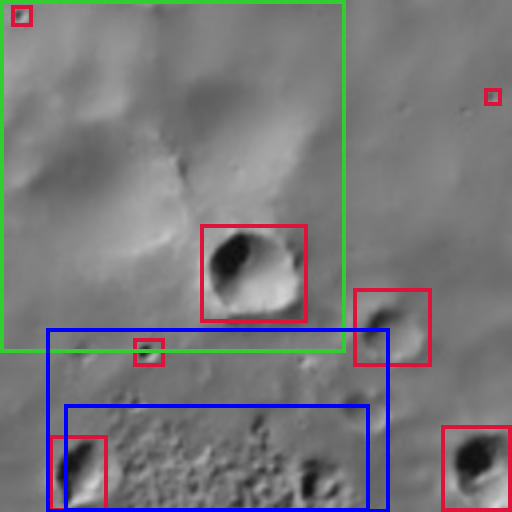}
        \caption{Source Only}
    \end{subfigure}
    % \hfil
    \begin{subfigure}{\qualsize\linewidth}
        \includegraphics{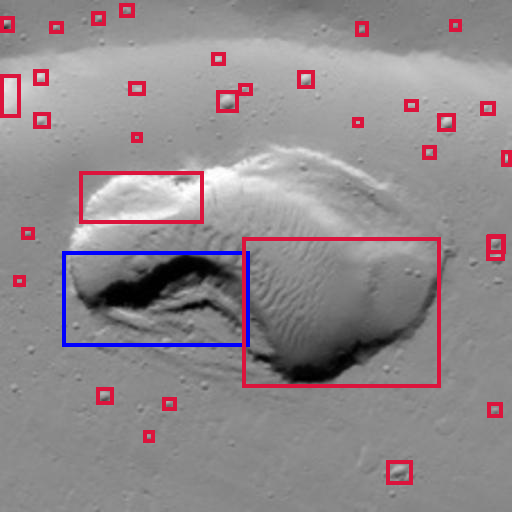}\\[3pt]
        \includegraphics{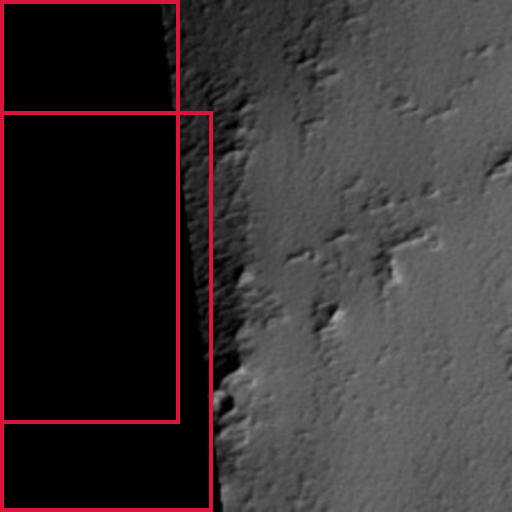}\\[3pt]
        \includegraphics{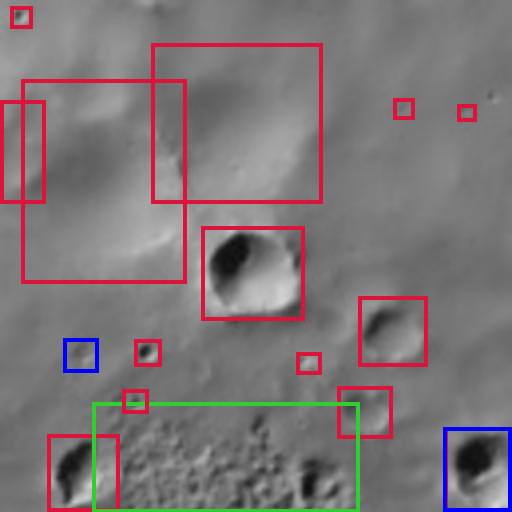}
        \caption{ViSGA~\cite{visga}}
    \end{subfigure}
    % \hfil 
    \begin{subfigure}{\qualsize\linewidth}
        \includegraphics{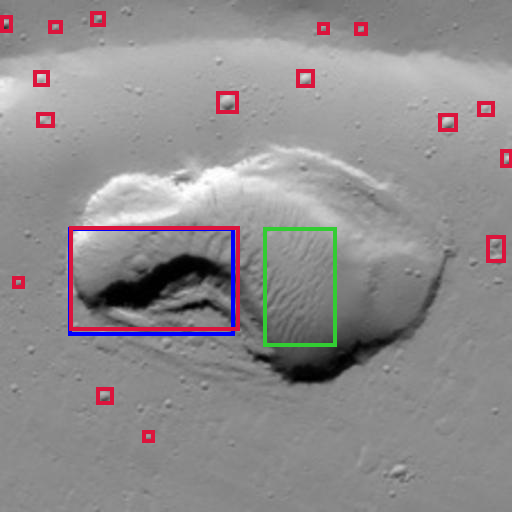}\\[3pt]
        \includegraphics{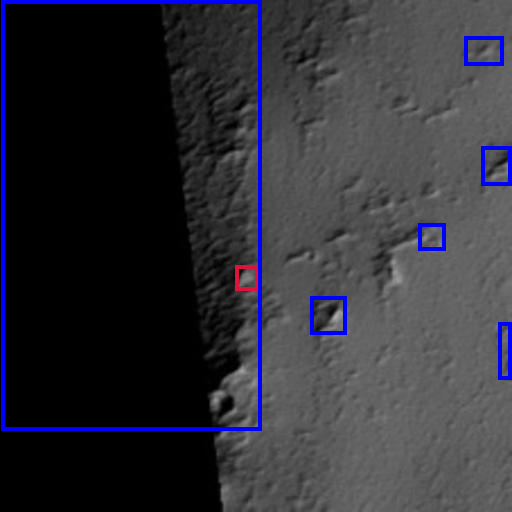}\\[3pt]
        \includegraphics{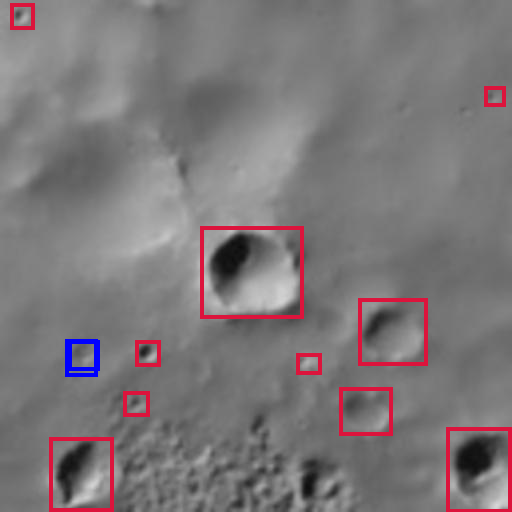}
        \caption{PC Only}
    \end{subfigure}
    % \hfil
    \begin{subfigure}{\qualsize\linewidth}
        \includegraphics{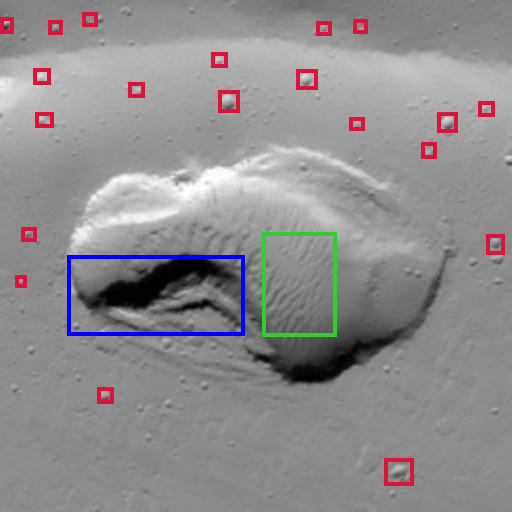}\\[3pt]
        \includegraphics{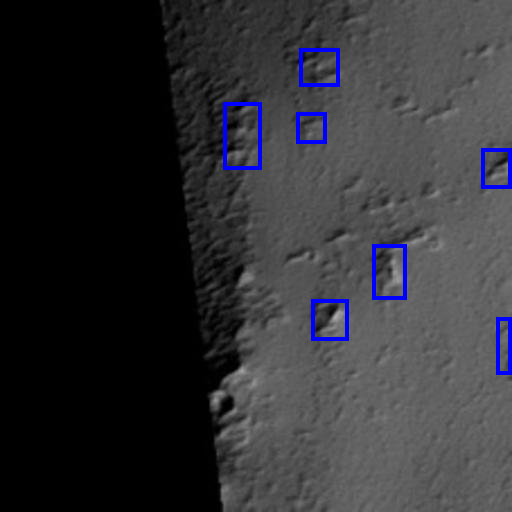}\\[3pt]
        \includegraphics{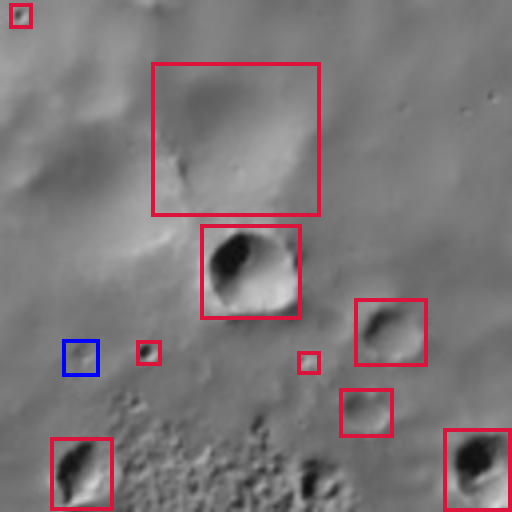}
        \caption{PC SFF}
    \end{subfigure}
    \caption{Qualitative detection examples on real-world Mars imagery~\cite{hirise} for \textit{Instance, Adversarial} techniques.}
    \label{fig:qual_mars}
    \vspace{-2em}
\end{figure*}

%% file: figs/qual/asteroid.tex
\begin{figure*}
    \centering
    \setkeys{Gin}{width=\linewidth}
    \begin{subfigure}{\qualsize\textwidth}
        \includegraphics{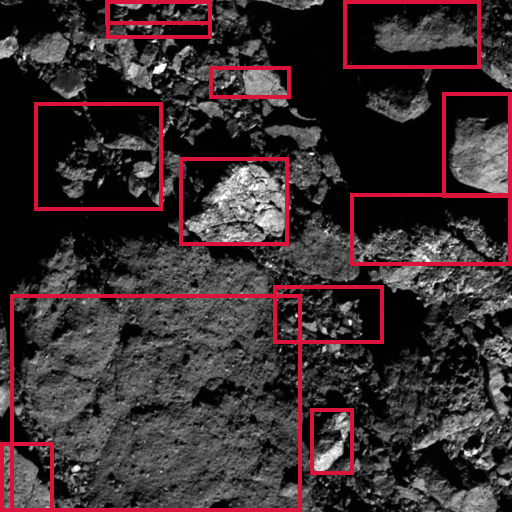}\\[3pt]
        \includegraphics{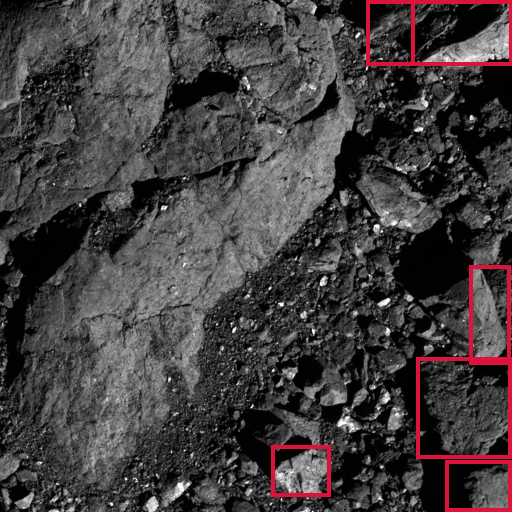}\\[3pt]
        \includegraphics{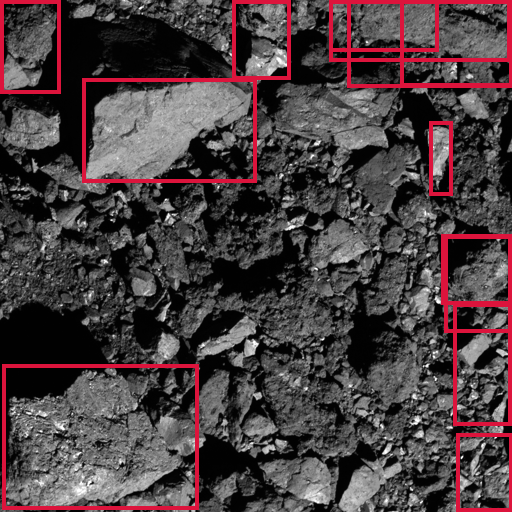}
        \caption{Source Only}
    \end{subfigure}
    % \hfil
    \begin{subfigure}{\qualsize\linewidth}
        \includegraphics{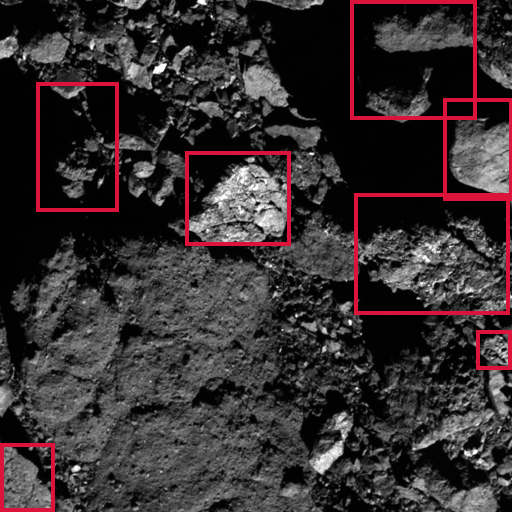}\\[3pt]
        \includegraphics{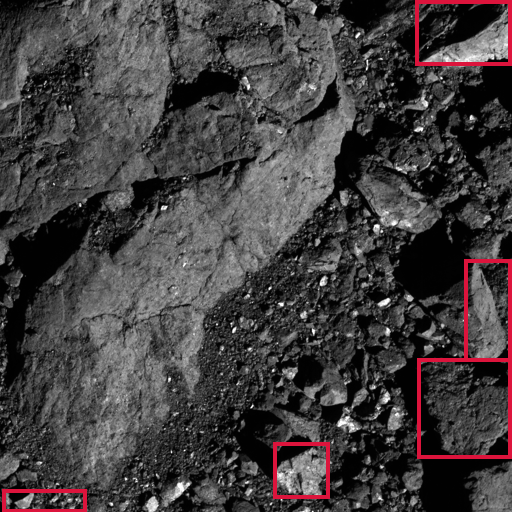}\\[3pt]
        \includegraphics{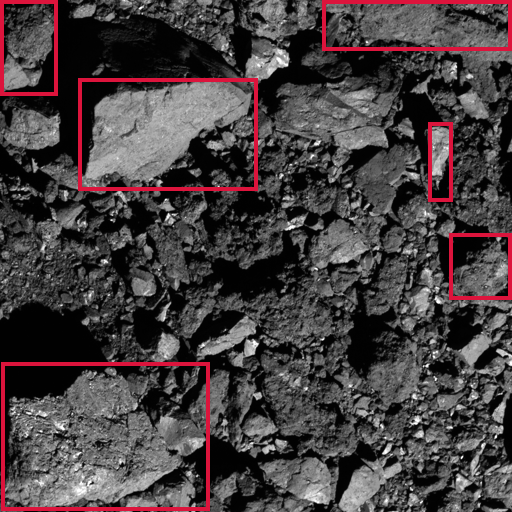}
        \caption{YOCOv1~\cite{yoco}}
    \end{subfigure}
    % \hfil
    \begin{subfigure}{\qualsize\linewidth}
        \includegraphics{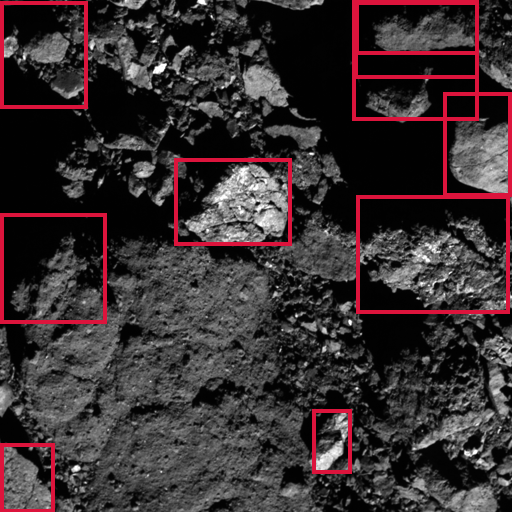}\\[3pt]
        \includegraphics{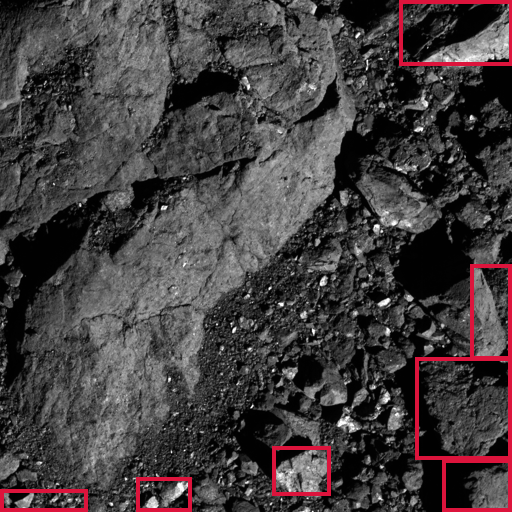}\\[3pt]
        \includegraphics{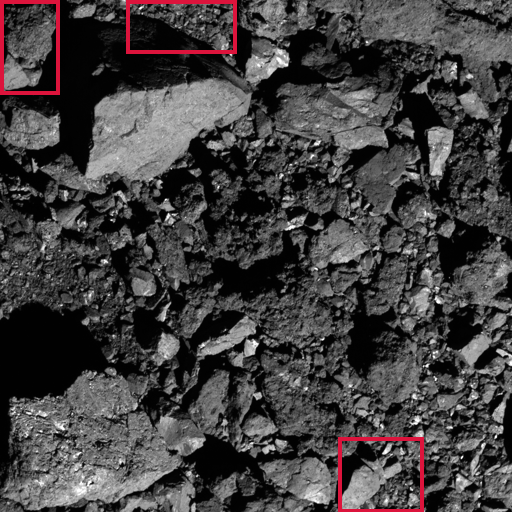}
        \caption{PC K-Means}
    \end{subfigure}
    % \hfil
    \begin{subfigure}{\qualsize\linewidth}
        \includegraphics{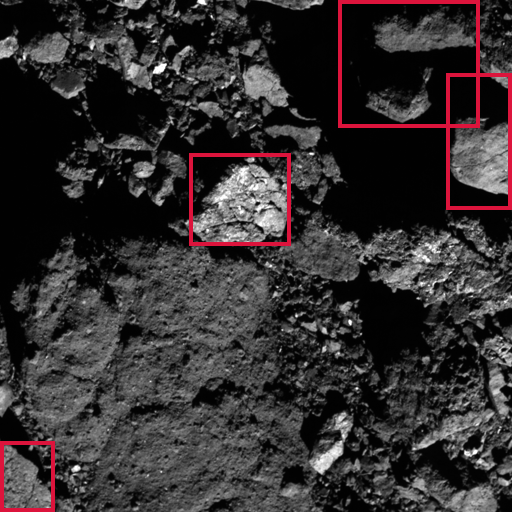}\\[3pt]
        \includegraphics{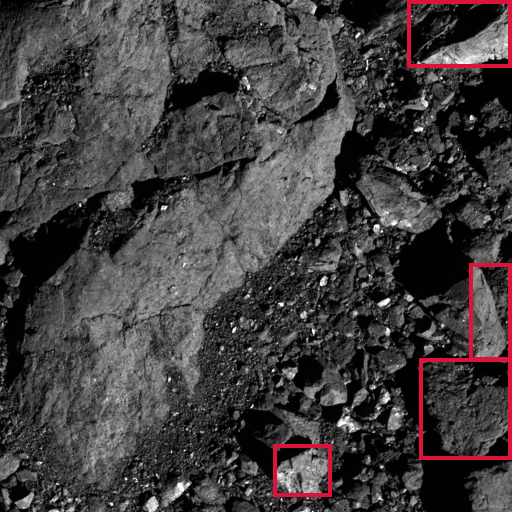}\\[3pt]
        \includegraphics{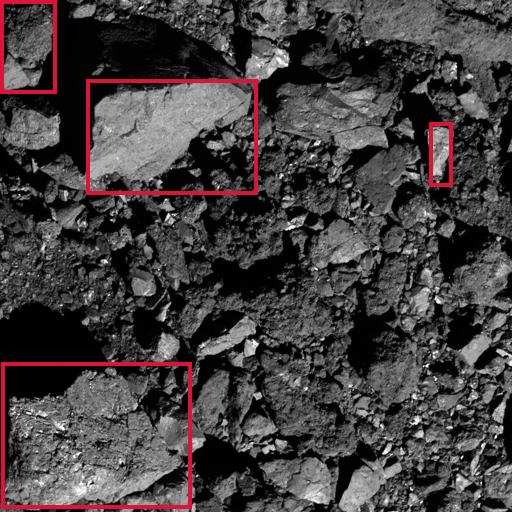}
        \caption{PTAP~\cite{ptap}}
    \end{subfigure}
    \caption{Qualitative detection examples on real-world Bennu asteroid imagery~\cite{orex_data} for \textit{Feature, Adversarial} techniques.}
    \label{fig:qual_asteroid}
    \vspace{-2em}
\end{figure*}

%% file: figs/qual/moon.tex
% \begin{figure*}
%     \centering
%     \subfigure[Source Only]{\includegraphics[width=0.249\textwidth]{figs/qual/moon/moon_2321_stock.png}}\hfill
%     \subfigure[Contrastive ROI]{\includegraphics[width=0.249\textwidth]{figs/qual/moon/moon_2321_roi.png}}\hfill
%     \subfigure[Contrastive Perceptual]{\includegraphics[width=0.249\textwidth]{figs/qual/moon/moon_2321_roi_fm.png}}\hfill
%      \subfigure[Contrastive Top-K]{\includegraphics[width=0.249\textwidth]{figs/qual/moon/moon_2321_roi_tk_fm.png}}
%     \caption{BLAH BLAH BLAH}
%     \label{fig:qual-moon}
% \end{figure*}

\begin{figure*}
    \centering
    \setkeys{Gin}{width=\linewidth}
    \begin{subfigure}{\qualsize\textwidth}
        \includegraphics{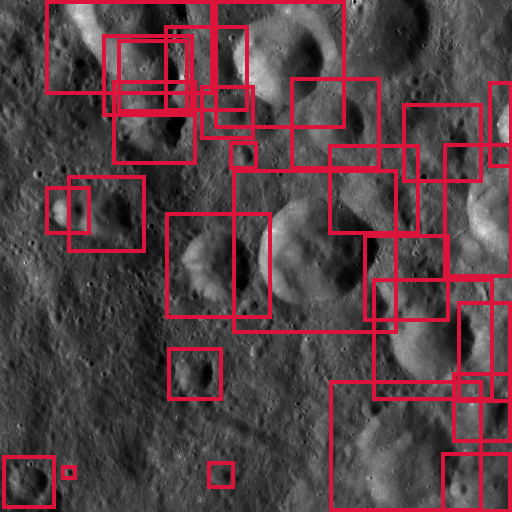}\\[3pt]
        \includegraphics{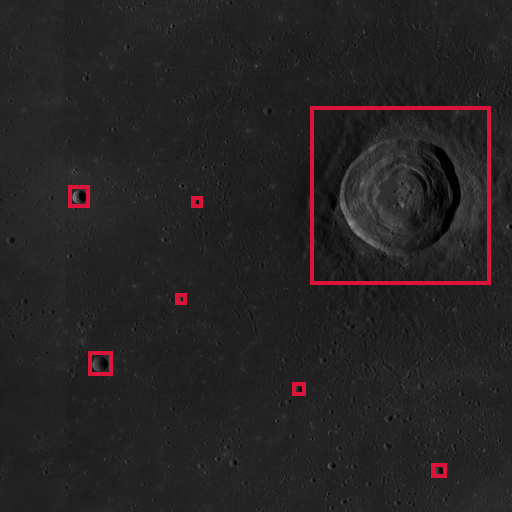}\\[3pt]
        \includegraphics{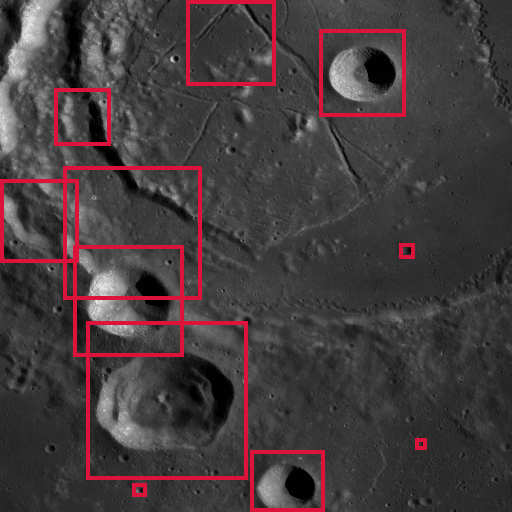}
        \caption{Source Only}
    \end{subfigure}
    % \hfil
    \begin{subfigure}{\qualsize\linewidth}
        \includegraphics{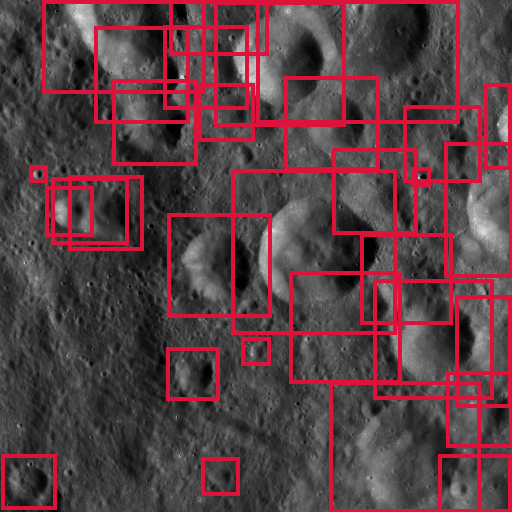}\\[3pt]
        \includegraphics{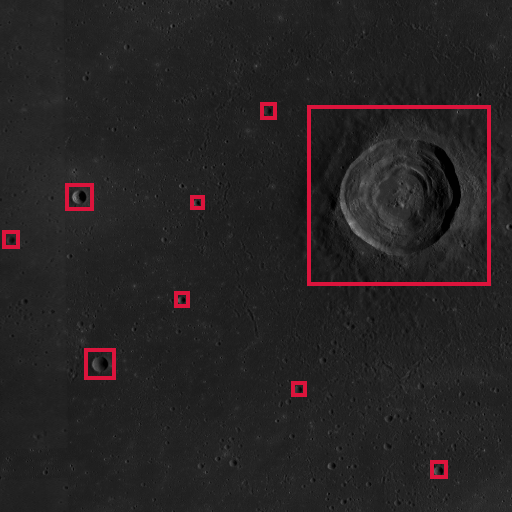}\\[3pt]
        \includegraphics{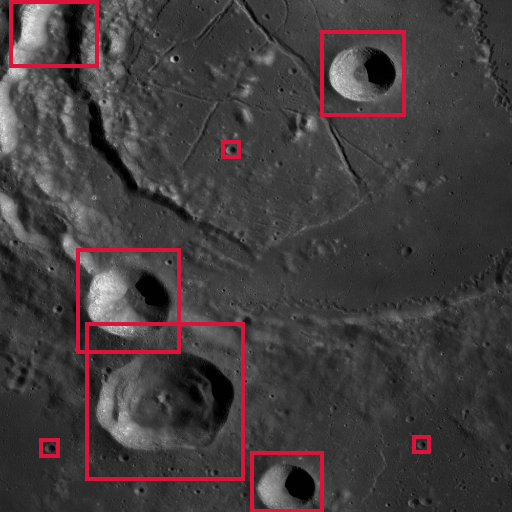}
        \caption{ViSGA~\cite{visga}}
    \end{subfigure}
    % \hfil
    \begin{subfigure}{\qualsize\linewidth}
        \includegraphics{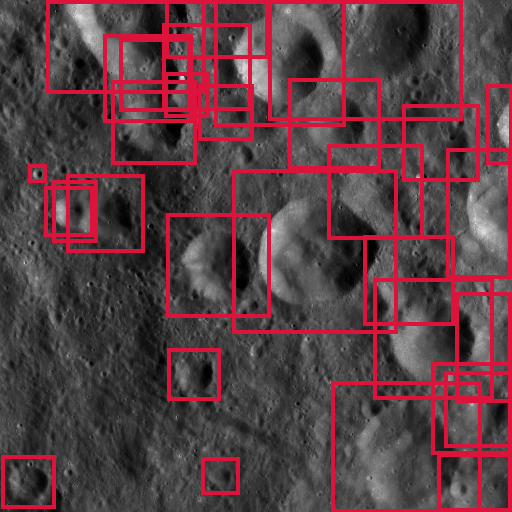}\\[3pt]
        \includegraphics{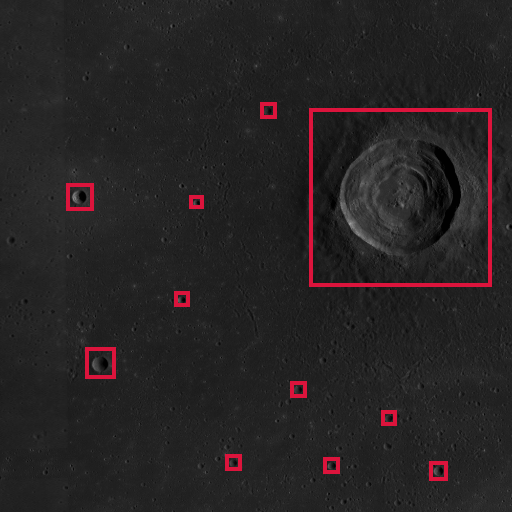}\\[3pt]
        \includegraphics{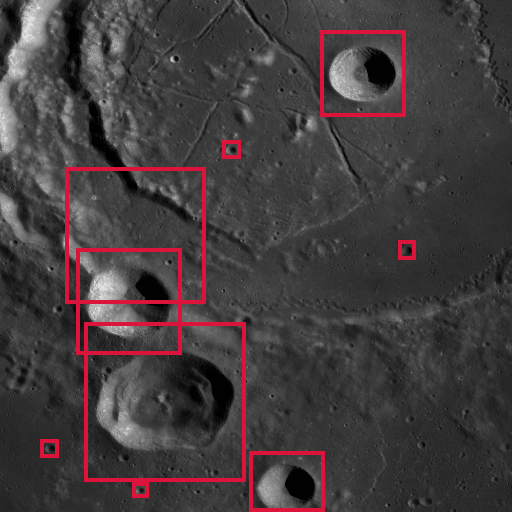}
        \caption{PC Only}
    \end{subfigure}
    % \hfil
    \begin{subfigure}{\qualsize\linewidth}
        \includegraphics{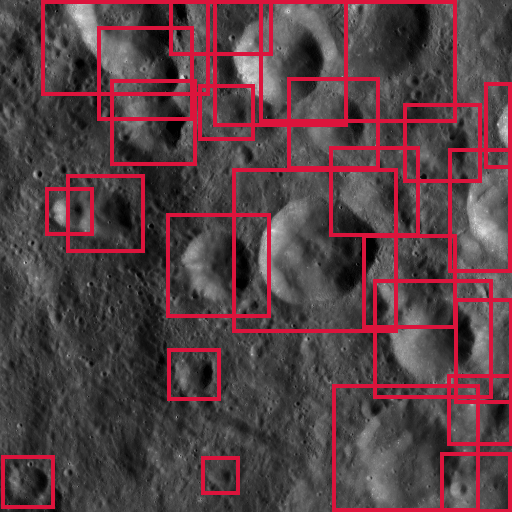}\\[3pt]
        \includegraphics{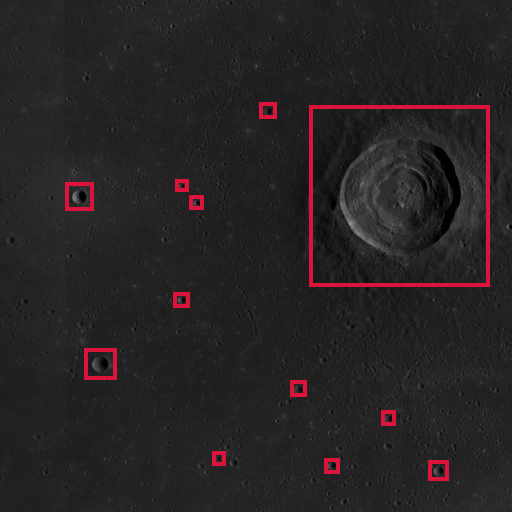}\\[3pt]
        \includegraphics{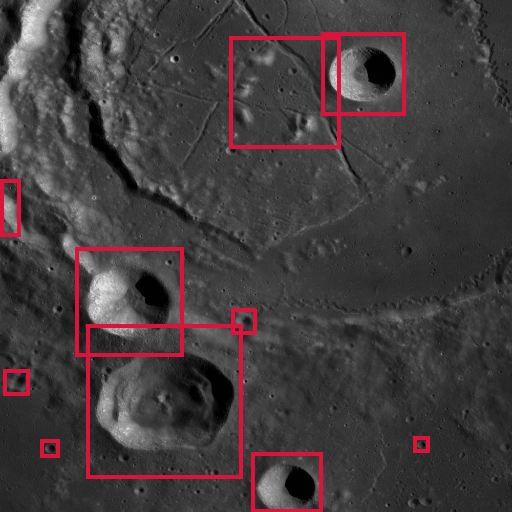}
        \caption{PC SFF}
    \end{subfigure}
    \caption{Qualitative detection examples on real-world Moon imagery~\cite{mosaic100} for \textit{Instance, Contrastive} techniques.}
    \label{fig:qual_moon}
    \vspace{-2em}
\end{figure*}

%% file: sections/conclusion.tex
\vspace{-1em}

\section{Conclusion}\label{sec:conclusion}
Novel techniques to Visual Similarity-based Alignment (VSA) for Unsupervised Domain Adaptation (UDA) in the context of one-stage, in-situ, and real-time terrain detection were proposed in this work. We demonstrate that the prior VSA implementation of intra-feature clustering found in YOCOv1 is largely insufficient for effective UDA performance with challenging space environments, and instrument a range of techniques that bolster the ability to learn domain-invariant properties of many classes of terrain. We performed an in-depth account of two techniques for general VSA, namely \textit{instance-based clustering} and \textit{intra-feature clustering}, and correlate the most prominent performance to the type of space environment, where we rigorously evaluate over Moon, Mars, and Asteroid scenes. Overall, our inception of perceptual consistency regularization and a robust feature selection mechanism ensures powerful domain transfer. We showcase quantitative performance on six datasets consisting of simulated and real-world data, observing VSA accuracy improvements upwards of 31\% compared to YOCOv1 and terrestrial state-of-the-art. Furthermore, we qualitatively examine detections on real-world mission imagery and see a reduction in erroneous terrain detections and improved accuracy.